\documentclass[10pt,twocolumn,letterpaper]{article}

\usepackage[pagenumbers]{cvpr} 

\usepackage{graphicx}
\usepackage{amsmath}
\usepackage{amssymb}
\usepackage{booktabs}
\usepackage{titletoc}
\usepackage{multirow}
\usepackage{capt-of}
\usepackage[table]{xcolor}
\usepackage[pagebackref,breaklinks,colorlinks]{hyperref}

\usepackage[capitalize]{cleveref}
\crefname{section}{Sec.}{Secs.}
\Crefname{section}{Section}{Sections}
\Crefname{table}{Table}{Tables}
\crefname{table}{Tab.}{Tabs.}

\definecolor{mylightblue}{rgb}{0.22, 0.45, 0.70} 

\hypersetup{
    colorlinks=true,
    linkcolor=mylightblue,
    citecolor=mylightblue,
    urlcolor=mylightblue
}
\def\cvprPaperID{XXXXX} 
\def\confName{CVPR}
\def\confYear{2026}

\begin{document}

\title{DuoGen: Towards General Purpose Interleaved Multimodal Generation}

\vspace{-0.3cm}
\author{
    Min Shi\textsuperscript{1,2,*},
    Xiaohui Zeng\textsuperscript{2,*},
    Jiannan Huang\textsuperscript{1},
    Yin Cui\textsuperscript{2},
    Francesco Ferroni\textsuperscript{2},
    Jialuo Li\textsuperscript{1}, \\
    Shubham Pachori\textsuperscript{2},
    Zhaoshuo Li\textsuperscript{2},
    Yogesh Balaji\textsuperscript{2},
    Haoxiang Wang\textsuperscript{2},
    Tsung-Yi Lin\textsuperscript{2}, \\
    Xiao Fu\textsuperscript{2},
    Yue Zhao\textsuperscript{2},
    Chieh-Yun Chen\textsuperscript{2},
    Ming-Yu Liu\textsuperscript{2,$\dagger$},
    Humphrey Shi\textsuperscript{1,2,$\dagger$}, \\ 
{\textsuperscript{1}Georgia Tech \quad \quad \textsuperscript{2}NVIDIA} \\
\href{https://research.nvidia.com/labs/dir/duogen/}{\texttt{research.nvidia.com/labs/dir/duogen}}
}

\twocolumn[{
    \maketitle
    \vspace{-25pt}
    \begin{center}
      \includegraphics[width=0.95\linewidth]{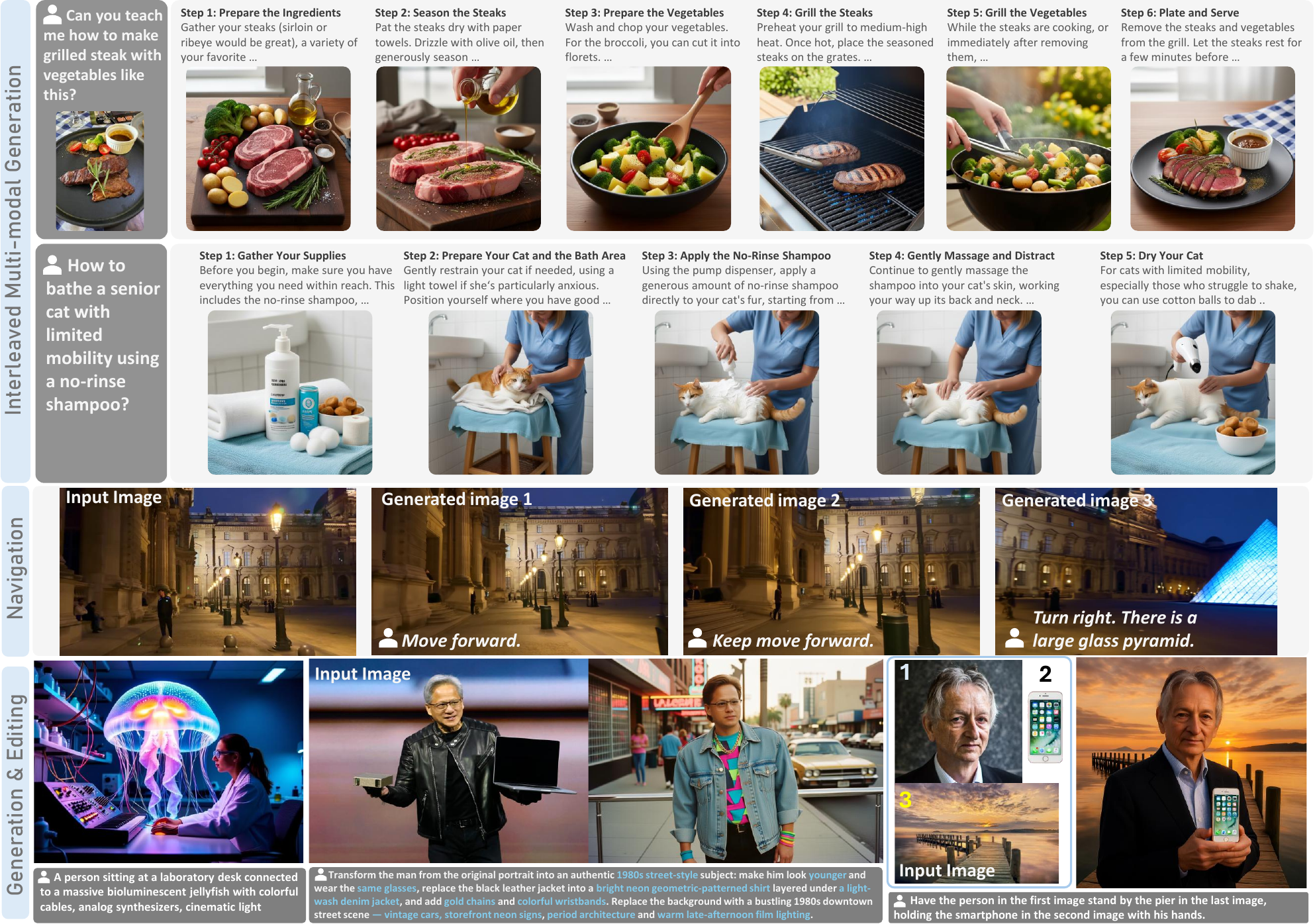}
      \vspace{-5pt}
      \captionof{figure}{
      \textbf{Capabilities of DuoGen.} Beyond standard tasks like image understanding, generation, editing, and navigation, DuoGen supports interleaved multimodal content generation, like step-by-step tutorials or cooking recipe. 
      }  
      \label{fig:fig-teaser}
    \end{center}
}]

\begingroup
  \renewcommand\thefootnote{}
  \footnotetext{
    $^*$Equal contribution
    \  
    $^\dagger$Corresponding authors
  }
\endgroup

\begin{abstract}
\vspace{-10pt}

Interleaved multimodal generation enables capabilities beyond unimodal generation models, such as step-by-step instructional guides, visual planning, and generating visual drafts for reasoning. However, the quality of existing interleaved generation models under general instructions remains limited by insufficient training data and base model capacity. We present DuoGen, a general-purpose interleaved generation framework that systematically addresses data curation, architecture design, and evaluation. On the data side, we build a large-scale, high-quality instruction-tuning dataset by combining multimodal conversations rewritten from curated raw websites, and diverse synthetic examples covering everyday scenarios. Architecturally, DuoGen leverages the strong visual understanding of a pretrained multimodal LLM and the visual generation capabilities of a diffusion transformer (DiT) pretrained on video generation, avoiding costly unimodal pretraining and enabling flexible base model selection. A two-stage decoupled strategy first instruction-tunes the MLLM, then aligns DiT with it using curated interleaved image–text sequences. Across public and newly proposed benchmarks, DuoGen outperforms prior open-source models in text quality, image fidelity, and image–context alignment, and also achieves state-of-the-art performance on text-to-image and image editing among unified generation models. Data and code will be released at \href{https://research.nvidia.com/labs/dir/duogen/}{Project Page}.

\end{abstract}

\section{Introduction}
\label{sec:intro}

Interleaved text–image generation enables a critical class of applications requiring tightly coupled multimodal outputs—such as step-by-step instructional guides, visual planning, and interactive editing—where text and visuals must be produced in a coordinated manner.
Although early works~\cite{gill,ge2023making,showo} show proof-of-concept results for storytelling or QA, they lack quantitative evaluation and are limited by their base models and data. 
Recent visual chain-of-thought systems~\cite{xu2025visual,li2025zebra,shi2025mathcanvas,gu2025thinkmorphemergentpropertiesmultimodal} generate images as visual drafts interleaved with textual thinking, but only in limited domains such as math and navigation. Despite these efforts, the field still lacks a systematic approach to general-purpose interleaved generation, spanning data, training, and evaluation. To fill this gap, we present DuoGen, a framework that holistically addresses all three components.

A major bottleneck for interleaved generation is the lack of high-quality, diverse instruction-tuning data, especially data with realistic user–assistant interactions. Although instruction tuning is essential for (multimodal) LLMs~\cite{flant5,instructgpt,llava}, existing efforts largely rely on large-scale interleaved pretraining corpora~\cite{mmc4,omnicorpus}, or video dense captions~\cite{deng2025emerging}. These sources provide limited instruction-style supervision. Recent visual chain-of-thought studies~\cite{li2025zebra,shi2025mathcanvas,gu2025thinkmorphemergentpropertiesmultimodal} interleave images with text as visual drafts for tasks like geometry or navigation, but they target reasoning rather than high-quality interleaved generation, and their task coverage remains narrow. To address the quantity, quality, and diversity gaps in instruction-tuning data, we curate 298k interleaved conversation samples from two complementary sources:
(1) a \textbf{data engine} that leverages a series of LLM/MLLM-based filtering and rewriting steps to convert raw webpages into clean user–assistant conversations; and (2) \textbf{synthetic data} generated by large language models and image generation models using carefully curated prompts designed to elicit high-quality images. 
For the data engine, we scrape 347k webpages from how-to sites, filtering the invalid webpages and images, then rewrite and convert the remaining passages into 268k conversations. The MLLM+LLM pipeline improves linguistic quality, enforces image–text coherence, and enables generating user inputs in arbitrary multimodal formats.
Though web-pages provide coherent real-world descriptions, their image aesthetics and resolutions are often limited due to lack of quality control. To enhance visual quality and consistency, we supplement the web data with 30k high-quality interleaved samples by large language models and image generation models. To ensure broad topic coverage, human annotators curate 1,500 seed prompts spanning 151 subcategories across 8 domains (\eg, home \& living, transportation), and we use OpenAI O3~\cite{o3-system-card-2025} to expand them into a diverse prompt pool. This curated synthetic subset substantially improves the visual quality of our instruction-tuning data.

To establish basic interleaved generation abilities, most unified models~\cite{team2024chameleon,showo} adopt an early-fusion paradigm that jointly trains on interleaved and unimodal generation tasks, such as text and images. Some works~\cite{li2025zebra,shi2025mathcanvas} also attempt to fine-tune from these pretrained models. However, unimodal pretraining requires heavy data engineering and computation, and restricts the choice of base models when scaling to different capacities. Recent unified systems~\cite{pan2025transfer,wu2025omnigen2,lin2025uniworld} combine pretrained image generators with MLLMs, but their interleaved generation remains underexplored or limited by architectural constraints. For example, the adopted image generation heads cannot accept multiple conditioning images.
This raises a key question: \textit{Can interleaved alignment be implemented directly on pretrained models without extensive unimodal pretraining?} Motivated by this question, we adopt a decoupled and scalable design that directly builds upon a pretrained MLLM and a diffusion transformer (DiT) pretrained on video generation. We name this framework DuoGen. DuoGen inherits the MLLM’s visual understanding and world knowledge to generate text, while the video-pretrained DiT enables generation of image sequences with consistent objects and scenes.
Concretely, the MLLM predicts a special token, \textless Begin-of-Vision\textgreater \ (BOV), to trigger image generation. To generate a new image, the previous images within the interleaved conversation history, either input or generated, are treated as conditioning frames for the DiT, while the MLLM hidden states preceding the \textless BOV\textgreater \ token provide semantic and linguistic guidance. This modular framework supports diverse choices of strong pretrained DiT and top-performing MLLMs without the need of unimodal pretraining from scratch and balancing understanding and generation objective in joint learning.

Together with the model design, we propose a two-stage decoupled training strategy that postpones interleaved pretraining while preserving the performance of the pretrained MLLM. In the first stage, we fine-tune only the MLLM using curated, high-quality interleaved generation data under next-token-prediction supervision. This stage teaches the MLLM to appropriately trigger image generation through \textless BOV\textgreater \ token and to continue text generation based on generated visuals. In the second stage, referred to as the interleaved context alignment stage, we freeze the MLLM parameters and update the DiT. Beyond the instruction-tuning data, this stage leverages large-scale interleaved alignment data, including interleaved image–text sequences that capture transitions between frames extracted from 5 million videos, as well as open-source image generation and editing samples.

We evaluate DuoGen on two public interleaved generation benchmarks: CoMM~\cite{chen2025comm} and InterleavedBench~\cite{interleavedbench}, which cover diverse tasks such as how-to questions and story generation, as well as different input formats (\textit{e.g.}, generation from scratch and continuation). In addition, we construct a new Interleaved Benchmark, focusing on diverse everyday problems. This benchmark leverages recent MLLMs capable of identifying fine-grained issues and includes the latest unified models such as NanoBanana~\cite{nanobanana} and Zebra-CoT~\cite{li2025zebra} fine-tuned from Bagel~\cite{deng2025emerging}.
Across all three benchmarks, DuoGen consistently outperforms previous open-source methods by a substantial margin across multiple metrics, including text quality, image fidelity, completeness, and image–context alignment. Moreover, DuoGen achieves significant gains on text-to-image and image-editing benchmarks compared to unified models like Bagel~\cite{deng2025emerging} and OmniGen2~\cite{wu2025omnigen2}, underscoring the benefits of leveraging pretrained MLLMs and video generation models. We will release both the model and dataset to facilitate future research on interleaved generation.

Our contribution can be summarized as follows:
\begin{itemize}
    \item We curate a high-quality 298k instruction-tuning dataset for interleaved generation, along with large-scale interleaved-alignment data.
    \vspace{-3pt}
    \item We design a model architecture that leverages strong unimodal generation models and introduce a novel, decoupled training strategy.
    \vspace{-3pt}
    \item We propose a benchmark for evaluating interleaved generation and provide comprehensive comparisons with existing open-source and commercial models.
\end{itemize}

\section{Related Work}
\noindent \textbf{Unified model.} Unified models aim to support both text and image generation within one model. Starting from Chameleon~\cite{team2024chameleon}, some works~\cite{vilau,liu2024lumina,chern2024anole,ge2023making} convert images into discrete tokens and unify language and text generation under next-token-prediction. Others, such as Transfusion~\cite{transfusion}, Bagel~\cite{deng2025emerging}, and the Show-o series~\cite{xie2025show,showo}, adopt a hybrid design that uses next-token prediction for text and diffusion for images. Another line of works use discrete-diffusion approaches to unify language and text generation, including MMaDA~\cite{yang2025mmada} and Lumina-DiMOO~\cite{xin2025lumina}.
In terms of training strategy, early-fusion models~\cite{team2024chameleon,transfusion,deng2025emerging} train from scratch on mixed text, images, and large-scale interleaved sequences, which requires substantial data and compute. In contrast, some works~\cite{pan2025transfer,lin2025uniworld,chen2025blip3} fuse a pretrained MLLM with a pretrained generator via different connector designs. Given the high cost of early-fusion training, we follow the pretrained-fusion approach while noting that our data, evaluation, and training strategies are also applicable to early-fusion pipelines.

\noindent \textbf{Interleaved generation model and datasets.}
Although unified models can generate both text and images, most still require users to specify the output modality and cannot seamlessly alternate between modalities to generate interleaved content. Early attempts~\cite{xie2025show,chern2024anole,ge2023making} demonstrate simple story-telling and how-to cases without quantitatively benchmarking these capabilities, and their output resolution remains limited. CoMM~\cite{chen2025comm} improves over noisy web-scale pretraining by converting how-to webpages into multimodal conversations. However, its data still contains stylistic noise (e.g., external links, inconsistent tone) and low-quality user-uploaded images, motivating the need for a more rigorous data pipeline. Visual chain-of-thought methods~\cite{shi2025mathcanvas,gu2025thinkmorphemergentpropertiesmultimodal,li2025zebra} further use generated images to assist reasoning, but their data focuses on several predefined tasks such as navigation or counting, limiting generalization ability. Based on these issues, we build a data engine that filters and rewrites web content using LLMs/MLLMs, and use high-quality synthetic data to improve visual fidelity and text-image alignment.

\section{Interleaved Multimodal Training Data}
The training data of DuoGen is divided into two parts: 1) high-quality interleaved multimodal conversations that teach models to follow user instructions; 2) interleaved image-text sequences for context alignment. 

\subsection{Instruction Tuning Data}
\label{sec:interleaved-instruction-tuning-data}

High-quality instruction-tuning data for interleaved generation remains extremely limited. To overcome both the quality and diversity constraints of existing data, we construct an interleaved instruction dataset from two complementary sources that jointly cover realistic, embodied, and visually high-fidelity cases.

\begin{figure}[t]
  \centering
  \includegraphics[width=0.7\linewidth]{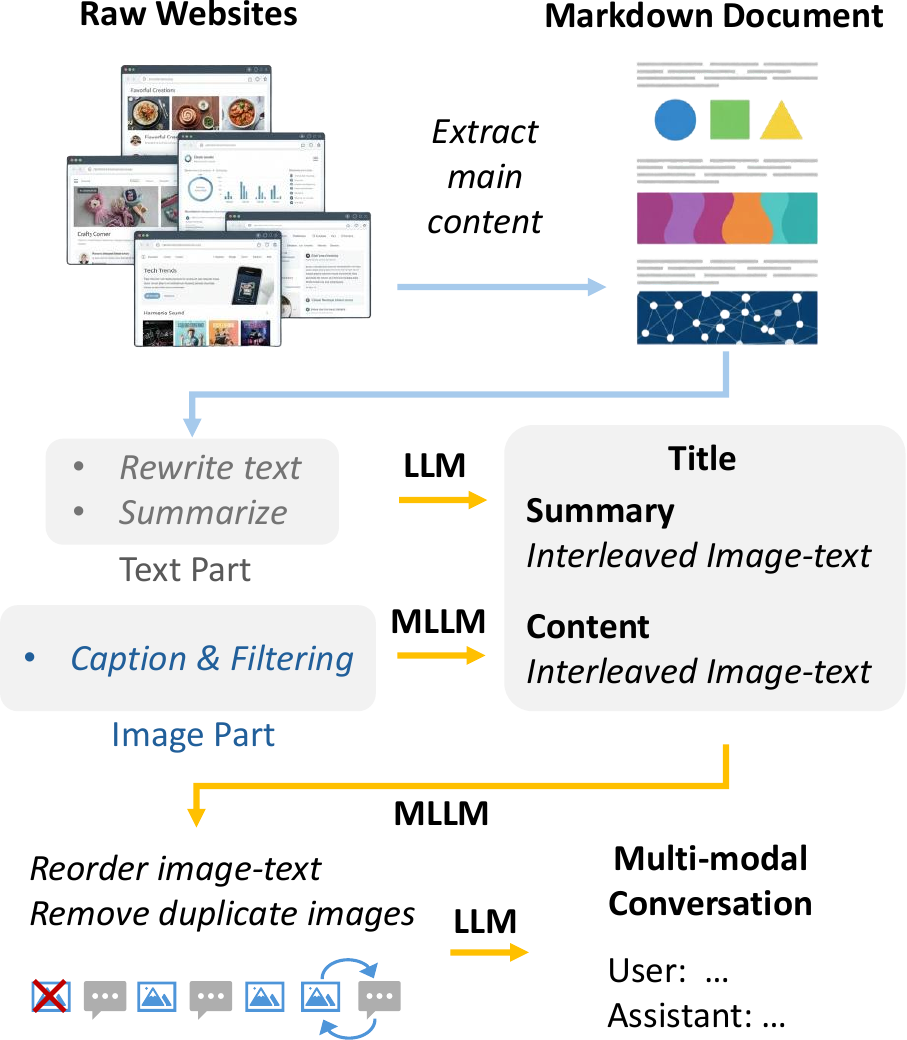}
   \caption{\textbf{Data engine for processing website data.} We design a data engine consists of a series of filtering and rewriting steps to convert noisy website data into high-quality instruction tuning data for interleaved generation.}  
   \label{fig:website-data-engine}
\end{figure}

\noindent \textbf{Data engine for websites.} The data engine converts raw webpages into multimodal conversations. Similar to CoMM~\cite{chen2025comm}, we source data from public how-to and story-telling websites, but introduce extensive post-processing and filtering, as illustrated in Fig.~\ref{fig:website-data-engine}. We collect webpages from StoryBird~\cite{storybird}, Instructables~\cite{instructables}, and eHow~\cite{ehow}, and also reuse available raw data from CoMM~\cite{chen2025comm} as an additional starting point. After removing pages containing only text or invalid images (e.g., QR codes, icons, advertisements), we retain 268k high-quality webpages out of 347k raw sites.. The main body of each webpage is converted into Markdown format for structured processing. 

Our pipeline consists of two major steps: (1) content rewriting and reorganization, and (2) conversion to user–assistant dialogue. First, we process text and images separately. Text passages are rewritten by an LLM to remove artifacts such as HTML tags, formatting errors, and external links. All images are captioned and categorized (e.g., natural photos, GUI screenshots, document pages), and invalid or irrelevant ones are discarded. To ensure coherence, we prompt an MLLM to remove duplicate or near-identical consecutive images and reorder image–text pairs so that each image appears after its corresponding description. Finally, a multimodal LLM transforms the cleaned image–text sequences into realistic instruction-style dialogues, where the user may optionally provide an image and the assistant responds step-by-step with interleaved reasoning and visual illustrations. In contrast to prior pipelines~\cite{mmc4,chen2025comm} without further rewriting and reorganization, our data engine actively denoises, restructures, and dialogizes web content, producing clean interleaved data for instruction tuning.

\noindent \textbf{High-quality synthetic data.} While website-derived data provide feasible real-world solutions, their image quality, resolution, and step granularity vary widely due to differences in user devices and content creation skills. The inconsistency can be harmful to the image quality of the generated interleaved sequence. To address these inconsistencies, we augment the webpage data with high-quality synthetic interleaved samples.

First, we need to prepare a pool of prompts covering different user queries. To enrich query diversity, we design a hierarchical query pool spanning eight broad everyday domains (e.g., Home \& Living, Pets \& Animal Caring). Domain annotators further refine these into 151 subcategories and compose about 10 seed questions per subcategory, yielding 1,500 seed prompts. Using OpenAI O3~\cite{o3-system-card-2025} with the highest reasoning budget, we expand these into 15,270 diverse instructions. During the expansion, the base categrory and other subcategories are also provided to avoid duplication. We then prompt the image generation model and image generation model to create the image.
In practice, we find this procedure performs particularly well on cooking-related tasks. We therefore additionally sample 15k dish images from MM-Food-100k~\cite{mmfood} as prompts for synthetic data generation. 

In total, we obtain around 30k prompts for high-quality synthetic data, reserving 700 prompts for evaluation. The website and synthetic data complement each other -- the synthetic portion provides high-resolution, stylistically consistent, and aesthetically appealing visuals that facilitate stable model learning.

\subsection{Interleaved Data For Context Alignment}
\label{sec:interleaved-alignment-data}

The interleaved data used for context alignment focuses on teaching the model to generate images consistent with preceding images and text. Unlike instruction-tuning data, these samples do not require meaningful linguistic interactions between a user and an assistant, making them relatively easy to acquire at scale. We leverage two primary sources: video transition captions and various image-generation tasks. For video data, following Bagel~\cite{deng2025emerging}, we collect 5 million raw videos and segment each into 5-second clips. All videos are pre-processed through scene detection and filtering to ensure temporal consistency within each segment. For every clip, we extract the first and last frame and annotate the transition using Qwen2.5-VL-32B~\cite{bai2025qwen2}, describing object motion, human actions, and camera movements. This converts raw videos into interleaved image–text sequences where the text explicitly explains the visual transition between frames. For image generation data, we aggregate open-source datasets including ShareGPT-4o-Image~\cite{chen2025sharegpt}, NHR-Edit~\cite{kuprashevich2025nohumansrequired}, OmniGen1\&2~\cite{omnigen,wu2025omnigen2}, UniWorld-V1~\cite{lin2025uniworld}, and Echo-4o~\cite{ye2025echo}, covering text-to-image, image editing, and multi-reference generation. Compared to video data, which typically captures smooth, subtle transitions, these datasets teach the model creative visual manipulation skills, such as adding, removing, or replacing objects and modifying backgrounds, which are also essential for general interleaved generation.

\begin{figure*}[t]
  \centering
  \includegraphics[width=1.0\linewidth]{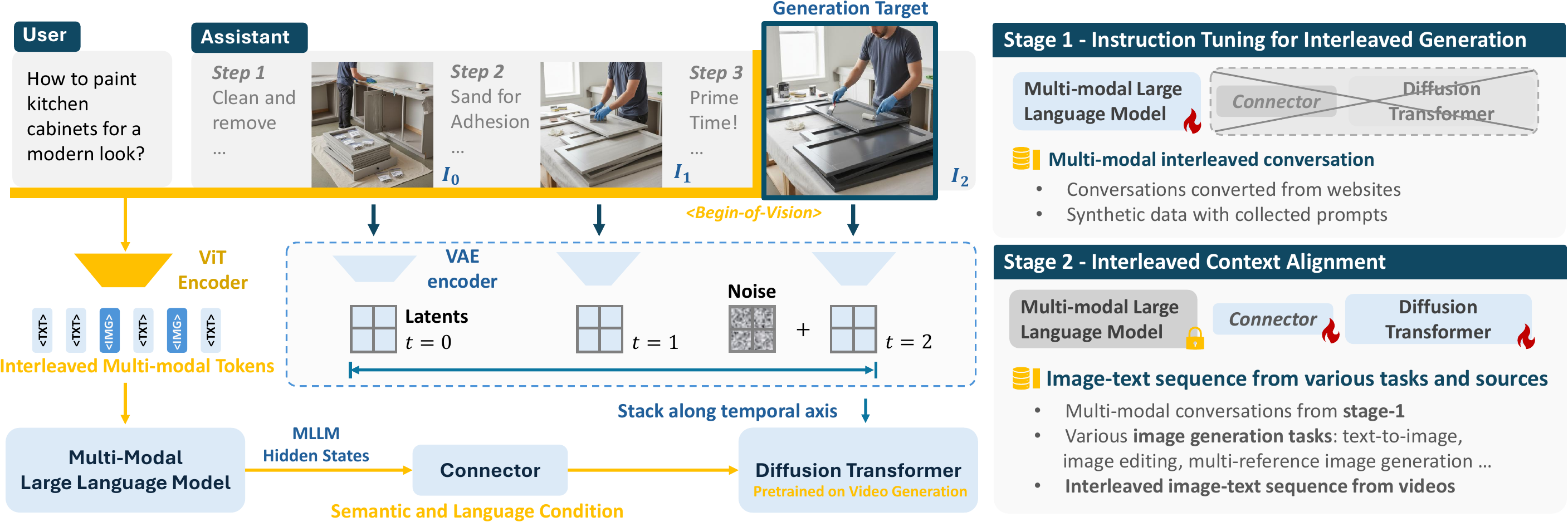}
   \caption{\textbf{Architecture and training strategy of DuoGen.} DuoGen consists of a pretrained multimodal large language model (MLLM) and diffusion transformer (DiT) pretrained on video generation. If a ``\textless Begin-of-Vision\textgreater'' (BOV) token is generated by the MLLM, then all the images in the interleaved sequence is packed as ``condition frames" to the DiT and the MLLM hidden-states before the \textless BOV\textgreater \ token is sent to the DiT as the text condition to generate the new images.}  
   \label{fig:model-architecture}
\end{figure*}

\section{Interleaved Generation Model}
In this section, we introduce the architecture and training strategy of DuoGen. Prior interleaved generation models, such as Show-o2~\cite{xie2025show} and Chameleon~\cite{team2024chameleon}, adopt an early-fusion paradigm that jointly pretrains unimodal and interleaved generation abilities, requiring substantial effort to build both image understanding and image generation capabilities from scratch. In contrast, modern pretrained MLLMs and video generation models already provide strong multimodal reasoning and high-quality visual generation. This raises a natural question: can we directly leverage these pretrained capabilities and enable interleaved generation on top? To answer this, we design a framework that fuses a pretrained MLLM with a pretrained video generation model. Under this formulation, the unified model only needs to learn two behaviors:
(1) the MLLM must autonomously trigger image generation when visual predictions benefit reasoning or user tasks, and
(2) the video generator must produce images consistent with prior text and images, whether user-provided or model-generated.

As shown in Fig.~\ref{fig:model-architecture}, DuoGen consists of an MLLM for text generation and a diffusion transformer (DiT) initialized from a video generation model for image synthesis. The MLLM can be any mainstream architecture equipped with a vision encoder and an LLM backbone, such as Qwen2.5-VL~\cite{bai2025qwen2} or LLaVA~\cite{llava}. The video generation component can be any model capable of conditioning on both images and text, such as Wan~\cite{wan2025wan} or the Cosmos-Predict series~\cite{agarwal2025cosmos}.

During generation, the MLLM autoregressively predicts the next token. When a special \texttt{<BOV>} token (Begin-of-Vision) is generated, the model is switched into image-generation mode. Once \texttt{<BOV>} is produced, assume the preceding interleaved sequence is ${T_1, I_1, T_2, I_2, \cdots, T_N}$, consisting of both user-provided and previously generated multimodal content. Then the DiT part needs to generate image ${I_N}$ conditioned on this sequence.
For the visual latent input, we stack all images appearing before the \texttt{<BOV>} token along the temporal axis to form a set of conditioning frames, and encode them into latent embeddings using the VAE encoder. These latents are concatenated with the noisy latent of the target image to construct the visual input to the video generator.
For the semantic and language condition, we extract the MLLM hidden states corresponding to all multimodal tokens preceding the \texttt{<BOV>} token. A lightweight connector projects these hidden states to the dimensionality required by the language-conditioning interface of the DiT.

During training, text generation is supervised with next-token prediction loss, masking out user input in the standard MLLM manner. The \texttt{<BOV>} token in the assistant turn is included in the loss, allowing the model to learn when to trigger image prediction. For image generation, we randomly sample one target image from each interleaved sequence, select a random diffusion step from the scheduler, and compute the loss (e.g., flow-matching~\cite{flowmatching}). During inference, the model autoregressively produces text until either a \texttt{<BOV>} token or the end-of-sequence token is reached. Once an image is generated, it is appended to the interleaved context, and the process repeats for subsequent steps. We further apply classifier-free guidance to enhance image fidelity: when generating the negative velocity, we keep the visual conditions fixed but remove the final text chunk from the MLLM hidden-state sequence.

\subsection{Implementation Details}
In this section, we introduce how to improve training efficiency. 
We adopt Qwen2.5-VL 7B~\cite{bai2025qwen2} architecture for the MLLM and initialize the DiT backbone using Cosmos Predict 2.5 (2B)~\cite{ali2025world}.

\noindent \textbf{Packed sequence training.} Sequence packing has become standard in MLLM/LLM training, allowing samples of different lengths and image resolutions to be packed together without padding and thereby improving training efficiency. However, the original implementation of Cosmos Predict 2.5~\cite{NvidiaCosmosPredict2} is incompatible with interleaved samples containing images of heterogeneous sizes.
To enable packed training, we introduce the following modifications: 1) For each interleaved sample, all images -- regardless of resolution -- are extracted and treated as a heterogeneous sequence of “video” frames. Their VAE latents are flattened and concatenated. For each image, we record its height, width, and index to restore the spatial shape during decoding; 2) We extend the original position embedding implementation. Now temporal indices increase by one after each image in the interleaved sequence, and the spatial RoPE (height/width indices) is computed using the per-image resolution.

\noindent \textbf{Condition input.} 
For text conditioning, guidance is injected via cross-attention between the image latents and the language embedding at every DiT decoder layer. Following Wang \textit{et al.}~\cite{llmt2i}, we concatenate the hidden states from all decoder layers along channel dimensiton to enhance representation. To prevent out-of-memory issues, we cap the maximum side length of images fed into the MLLM at 480 pixels. For visual conditioning, Cosmos Predict 2.5 concatenates the condition image latents with the noisy latents of the target frame along the temporal axis. We adopt the same strategy: the clean latents of user-provided images and previously generated images are concatenated with the noisy latent corresponding to the current generation target, forming a unified visual condition sequence.

\subsection{Decoupled Training Strategy}
Based on our interleaved data and model architecture, we adopt a decoupled two-stage training strategy. As illustrated in Fig.~\ref{fig:model-architecture}, training is divided into: (1) instruction tuning of the MLLM for interleaved generation, and (2) interleaved context alignment for the connector and DiT.
In the first stage, we update only the MLLM parameters using the high-quality multimodal conversations described in Sec.~\ref{sec:interleaved-instruction-tuning-data}, supervised with next-token prediction. After this stage, the MLLM learns to autonomously trigger image generation at appropriate moments and to continue text generation conditioned on newly produced images. We intentionally exclude data for interleaved context alignment discussed in Sec.~\ref{sec:interleaved-alignment-data} here, as such data lacks meaningful user–assistant interactions; introducing it too early may harm the pretrained MLLM’s carefully engineered post-training behaviors.
In the second stage, we freeze the MLLM and fine-tune only the connector and DiT, using the context-alignment data from Sec.~\ref{sec:interleaved-alignment-data}, which includes video-labeled interleaved sequences and diverse image generation/editing datasets. We also add the instruction tuning data in Sec.~\ref{sec:interleaved-instruction-tuning-data} into the training. This enables image generation that stays well aligned with preceding images and textual context.
This decoupled approach also let us leverage heterogeneous data effectively: even if text from the alignment data may be uninformative for a strong pretrained MLLM, it remains valuable for aligning visual generation behavior. The same strategy is applicable to other unified frameworks with separated language and diffusion parameters, such as Bagel~\cite{deng2025emerging}.

\section{Experiment}
In this section, we present results on interleaved generation benchmarks, followed by evaluations on image generation and image editing tasks. We additionally conduct ablation studies on different data recipe to validate the effectiveness of our data engine and the contribution of synthetic interleaved data.

\begin{table*}[!t]
    \center
    \small
    \caption{\textbf{Comparison on interleaved generation tasks.} T-Com, I-Com, I-Co, IT-Co, I-Q denotes text completeness, image completeness, image-coherence, image-text coherence, and image quality, respectively. 7B/2B in size column denotes the activated parameters for text generation and image generation if using decoupled design.}
    \addtolength{\tabcolsep}{-3.5pt}
    \begin{tabular}{lc|ccccc|ccccc|ccccc}
    \toprule
    \multirow{2}{*}{Model} & \multirow{2}{*}{Size} & \multicolumn{5}{c|}{Cooking-200}      & \multicolumn{5}{c|}{Cooking-200-Text-Input} & \multicolumn{5}{c}{How-to-500}        \\
                           &                       & T-Com & I-Com & I-Co & I-Q & IT-Co  & T-Com   & I-Com  & I-Co  & I-Q  & IT-Co   & T-Com & I-Com & I-Co & I-Q & IT-Co  \\ \midrule
    {\color{gray}Nano Banana\cite{nanobanana}}             & {\color{gray}-}                     & {\color{gray}4.24}   & {\color{gray}4.07}   & {\color{gray}4.36} & {\color{gray}4.81}  & {\color{gray}4.83} & {\color{gray}4.02}     & {\color{gray}4.02}    & {\color{gray}4.59}  & {\color{gray}4.75}   & {\color{gray}4.72}  & {\color{gray}3.95}   & {\color{gray}4.28}   & {\color{gray}4.49} & {\color{gray}4.22}  & {\color{gray}4.24} \\ \midrule
    SEED-LLaMA\cite{ge2023making}             & 7B/0.8B                    & 1.99& 1.63& 2.93& 3.14& 1.65& 2.08     & 1.70    & 2.99  & 3.35   & 1.86  & 1.61   & 1.50   & 3.18 & 2.97  & 1.69 \\
    MiniGPT-5\cite{zheng2023minigpt}              & 7B/0.8B                      & 1.85& 1.81& 1.75& 2.81& 1.88& 2.03     & 2.27    & 1.84  & 2.91   & 2.10  & 1.94   & 2.22   & 2.63 & 2.98  & 2.43 \\
    Zebra-CoT\cite{li2025zebra}              & 7B/7B                 & 2.10   & 2.63   & 3.54 & 3.61  & 3.67 & 2.12& 2.54& 3.21& 3.06& 3.07& 2.04   & 2.05   & 3.52 & 2.84  & 2.59 \\ \midrule
    DuoGen                & 7B/2B                 & \textbf{3.61}   & \textbf{4.70}   & \textbf{3.92} & \textbf{4.78}  & \textbf{4.75} & \textbf{3.82}& \textbf{4.77}& \textbf{4.17}& \textbf{4.79}& \textbf{4.76}& \textbf{3.39}   & \textbf{4.22}   & \textbf{4.21} & \textbf{4.08}  & \textbf{4.18} \\ \bottomrule
    \end{tabular}
    \label{tab:comparison-our-interleaved-generation}
\end{table*}

\begin{table}[!t]
    \center
    \small
    \caption{\textbf{Comparison on CoMM~\cite{chen2025comm}}. Sty. and Enti. denotes the style and entity consistency among generated images. Tren. denotes the trend alignment betwen image and text squence. Comp. denotes the completeness, ImgQ is the image quality. IRS is the illustration relevance score which is used to measure whether the generated images fits the surrounding context. }
    \addtolength{\tabcolsep}{-2.5pt}
    \begin{tabular}{l|cccccc}
    \hline
    Model      & Sty. & Enti. & Tren. & Comp. & ImgQ & IRS  \\ \hline
    MiniGPT-5\cite{zheng2023minigpt}  & 5.65  & 5.2    & 5.25  & 5.81         & 6.15          & 2.71 \\
    SEED-LLaMA\cite{ge2023making} & 7.55  & 6.81   & 6.15  & 5.13         & 6.36          & 1.46 \\
    Emu2\cite{sun2024generative}       & 8.41  & 7.56   & 7.63  & 7.54         & 7.59          & 2.02 \\ \midrule
    DuoGen    & \textbf{9.22}  & \textbf{9.22}   & \textbf{9.24}  & \textbf{9.66}         & \textbf{9.53}          & \textbf{7.76} \\ \hline
    \end{tabular}
    \label{tab:comparison-comm-test}
\end{table}

\begin{table}[!t]
    \center
    \small
    \caption{\textbf{Comparison on InterleavedBench.} T-Q, I-Q, I-Co, IT-Co denotes text-quality, image-quality, image-coherence and the image-text coherence, respectively.}
    \addtolength{\tabcolsep}{-2.5pt}
    \begin{tabular}{l|cccccc}
    \hline
    Model     & T-Q & I-Q & I-Co & IT-Co & Helpfulness & Avg. \\ \hline
    MiniGPT-5\cite{zheng2023minigpt} & 1.22         & 2.45               & 1.62            & 2.03                 & 1.77        & 1.82 \\
    GILL\cite{gill}      & 0.75         & 3.21               & 2.25            & 1.53                 & 1.48        & 1.84 \\
    Emu2\cite{sun2024generative}      & 1.26         & 2.28               & 1.89            & 1.34                 & 1.64        & 1.68 \\ \midrule
    DuoGen   & \textbf{4.28}         & \textbf{3.65}               & \textbf{3.70}            & \textbf{3.69}                 & \textbf{4.06}        & \textbf{3.87} \\ \hline
    \end{tabular}
    \label{tab:comparison-interleavedbench}
\end{table}

\subsection{Interleaved Generation}
We evaluate interleaved generation on our benchmark and two public benchmarks: CoMM~\cite{chen2025comm} and InterleavedBench~\cite{interleavedbench}. CoMM~\cite{chen2025comm} contains text-only instructions covering story generation and how-to questions. InterleavedBench extends this setting with more tasks like passage generation and additional input formats like continuation tasks where models complete partially provided interleaved contexts. Both benchmarks rely on GPT-4o~\cite{openai-gpt4o-2024} for evaluation, scoring text and image completeness, image coherence, and image–text alignment. However, we observe that GPT-4o often misses fine-grained visual artifacts or subtle mismatches between user context and generated visuals.

To better evaluate results rigorously, we introduce a new benchmark, focusing on diverse, realistic tasks and employing stronger VLMs for evaluation. It includes two subsets: Cooking-200, where the model generates interleaved recipes consistent with a provided dish image and title, and How-to-500, an open-ended set of 500 everyday questions spanning 151 subcategories. Using an MLLM-as-judge protocol, we find that GPT-5~\cite{openai-gpt5-2025} reliably identifies subtle visual–semantic inconsistencies—for example, penalizing cases where the model generates headless shrimp despite the input image clearly showing shrimp with heads—while GPT-4o often overlooks such fine-grained errors. We report both sequence-level metrics (text completeness, image completeness, image coherence) and image-level metrics (aesthetic quality, image–text coherence). Refer to Supp.~\ref{supp-part:interleaved-benchmark-details} for more details.

\noindent \textbf{Quantitative Comparison.} Tables~\ref{tab:comparison-comm-test} and \ref{tab:comparison-interleavedbench} present results on CoMM~\cite{chen2025comm} and InterleavedBench~\cite{interleavedbench}. DuoGen consistently outperforms prior systems across all major dimensions, including text quality, image quality, visual coherence, and image–text alignment. On the CoMM test set, it achieves a substantial improvement in Illustration Relevance Score (IRS) measuring image-text alignment, reaching 2.8× the score of the second-best method (7.76 vs. 2.71 for MiniGPT-5). A similar trend is observed on InterleavedBench with continuation tasks that require interpreting user-provided images and contextual inputs, where DuoGen shows an even larger advantage in text quality, attaining 3.4× the score of Emu2. These results show that DuoGen can comprehend complex user inputs to generate coherent and helpful textual solutions, and produce high-quality images that remain closely aligned with the accompanying text, demonstrating the advantages of utilizing well-pretrained models. 

\begin{table*}[!t]
\centering
\footnotesize 
\begin{minipage}[t]{0.46\textwidth}
    \centering 
    \setlength{\tabcolsep}{3pt} 
    \caption{\textbf{Comparison on GenEval.} \small * denotes LLM prompt rewriting.** uses interleaved generation to improve image quality.}
    \begin{tabular}{c|lc}
    \toprule
    Model Type            & Method        & Overall \\ \midrule
    {\color{gray} Commercial}            & {\color{gray}GPT-4o-Image\cite{openai-gpt4o-2024}}  & {\color{gray}0.84}    \\ \midrule
    \multirow{4}{*}{Generation} & SDXL\cite{podell2023sdxl}          & 0.55    \\
                          & DALLE-3\cite{openai_dalle3_2023}       & 0.67    \\
                          & FLUX.1-dev\cite{flux2024}    & 0.82    \\
                          & Qwen-Image\cite{wu2025qwen}    & 0.87    \\ \midrule
    \multirow{9}{*}{Unified Model} & Emu3\cite{wang2024emu3}          & 0.54    \\
                          & Show-o\cite{showo}        & 0.53    \\
                          & Janus-Pro-7B\cite{chen2025janus}  & 0.80     \\
                          & MMaDA\cite{yang2025mmada}         & 0.63    \\
                          & MetaQuery-XL*\cite{pan2025transfer} & 0.80     \\
                          & Blip-3o\cite{chen2025blip3}       & 0.84    \\
                          & Bagel\cite{deng2025emerging}         & 0.82    \\
                          & UniWorld-V1\cite{lin2025uniworld}   & 0.80     \\
                          & OmniGen2\cite{wu2025omnigen2}      & 0.80     \\ \midrule
    \multirow{2}{*}{\begin{tabular}[c]{@{}c@{}}Interleaved \\  Generation\end{tabular}} & Uni-CoT**\cite{qin2025unicot} & 0.83    \\
                          & DuoGen          & 0.88    \\ \bottomrule
    \end{tabular}
    \label{tab:comparison-t2i-geneval}
\end{minipage}
\hfill
\begin{minipage}[t]{0.53\textwidth}
    \centering
    \setlength{\tabcolsep}{3pt} 
    \caption{\textbf{Combined comparison on ImgEdit and GEdit\_EN.} G\_SC, G\_PQ, and G\_O are sub-metrics for GEdit\_EN.}
    \begin{tabular}{c|l|c|ccc}
    \toprule
\multirow{2}{*}{Model Type} & \multirow{2}{*}{Method} & \multicolumn{1}{c|}{ImgEdit} & \multicolumn{3}{c}{GEdit\_EN} \\[3pt]
\multicolumn{1}{c|}{} & \multicolumn{1}{l|}{} & \multicolumn{1}{c|}{Overall} & G\_SC & G\_PQ & G\_O \\
    \midrule
    \multirow{2}{*}{\color{gray}Commercial}  & {\color{gray}Nano Banana\cite{nanobanana}} & {\color{gray}4.23} & {\color{gray}7.28} & {\color{gray}7.83} & {\color{gray}6.93} \\
      & {\color{gray}GPT-4o-Image\cite{openai-gpt4o-2024}} & {\color{gray}4.20} & {\color{gray}7.85} & {\color{gray}7.62} & {\color{gray}7.53} \\
    \midrule
    \multirow{4}{*}{Generation} & ICEdit\cite{zhang2025context} & 3.05 & 5.11 & 6.85 & 4.84 \\
     & Step1X-Edit\cite{liu2025step1x} & 3.06 & 7.09 & 6.76 & 6.701 \\
     & FLUX.1 Kontext {[}Pro{]}\cite{labs2025flux} & 4.00 & 7.02 & 7.6 & 6.56 \\
     & Qwen-Image-Edit\cite{wu2025qwen} & 4.27 & 8.00 & 7.86 & 7.56 \\
    \midrule
    \multirow{6}{*}{Unified Model} & OmniGen\cite{omnigen} & 2.96 & 5.96 & 5.89 & 5.06 \\
     & Bagel\cite{deng2025emerging} & 3.20 & 7.36 & 6.83 & 6.52 \\
     & UniWorld-V1\cite{lin2025uniworld} & 3.26 & 4.93 & 7.43 & 4.85 \\
     & OmniGen2\cite{wu2025omnigen2} & 3.44 & 7.16 & 6.77 & 6.41 \\
     & OVIS-U1\cite{wang2025ovis} & 4.00 & - & - & 6.42 \\
    \midrule
    \multirow{2}{*}{\begin{tabular}[c]{@{}c@{}} Interleaved \\ Generation \end{tabular}} & Uni-CoT\cite{qin2025unicot} & - & 7.91 & 6.24 & 6.74 \\
     & DuoGen & 4.19 & 7.68 & 7.76 & 7.35 \\
    \bottomrule
    \end{tabular}
    \label{tab:comparison-editing-combined}
\end{minipage}
\end{table*}

Table~\ref{tab:comparison-our-interleaved-generation} reports results on the two subsets of our benchmarks. 
Cooking-200-Text-Input removes images from user input. 
Nano Banana~\cite{nanobanana} shows strong performance across all the subsets, especially on How-to-500, which requires broader knowledge and the ability to generate physically plausible objects and procedures. DuoGen surpasses all other open-source models by large margins across all metrics, with particularly notable gains on How-to-500. Moreover, DuoGen significantly narrows the gap between open-source models and Nano Banana; on the more constrained Cooking-200 tasks, DuoGen even matches Nano Banana on certain metrics such as image–text coherence. These results highlight the potential of our framework: with sufficient high-quality data, DuoGen can approach the performance of top commercial models on specific domains.

\noindent \textbf{Qualitative Results.} Fig.~\ref{fig:fig-teaser} presents two interleaved generation examples. The model produces high-resolution images (768×768) with fine visual details and strong consistency both across generated frames and between user inputs and model outputs. In the grilled-steak example, DuoGen idenitify and generate the sides such as tomatoes, broccoli, and potatoes. In the bathe-a-cat example, the model maintains consistency of major objects—including the bathroom environment, human, and the cat—across multiple steps, demonstrating robust spatial and semantic coherence during interleaved reasoning and generation. Additional examples are provided in Supp.~\ref{supp-part:interleaved-generation-examples}.

\subsection{Image Generation and Editing}
We use GenEval~\cite{geneval} to evaluate the image generation capabilities and use ImgEdit~\cite{ye2025imgedit} and the English subset of  GEdit~\cite{liu2025step1x} to evaluate the image editing performance.

\noindent \textbf{Image Generation}. Our method significantly outperforms other unified generation models on the overall score. In particular, DuoGen achieves strong improvements on multi-object metrics such as counting (0.94), position (0.84), and attribute binding (0.80)—areas where unified models typically struggle—indicating enhanced compositional reasoning and spatial grounding. Overall, our approach substantially narrows the gap with state-of-the-art commercial and task-specialized generative systems, while establishing a new performance baseline for unified multimodal generation. See Supp.~\ref{supp-part:image-generation-editing-examples} for detailed results.

\noindent \textbf{Image Editing.} Table~\ref{tab:comparison-editing-combined} shows the result on ImgEdit~\cite{ye2025imgedit} and GEdit\_EN~\cite{liu2025step1x} benchmarks, which uses VLM to evaluate editing results from prompt following and visual quality. On ImgEdit~\cite{ye2025imgedit}, DuoGen significantly outperforms prior unified models, especially on more complex tasks like ``hybrid'', ``add'', and ``replace''. While recent editing model, Qwen-Image-Edit, still achieves the highest overall score, DuoGen is narrowing the gap as a interleaved generation model, and shows better score on Remove (4.71), Replace (4.69), and Add (4.53), demonstrating strong capability in precise object-level transformations. GEdit\_EN benchmark uses two metrics, ``G\_SC'' is for semantic consistency which evaluate whether editing is consistent with user prompt and ``G\_PQ'' is pixel quality. ``G\_O'' is the geometry average of ``G\_SC'' and ``G\_PQ''. Our model achieves strong performance across the three metrics compared with other unified model and closely matching on commercial models and strong editing models. The results on image generation and editing demonstrating the advantage of building upon DiT well-pretrained on video generation, which offers good pixel generation quality and content creation abilities.

\noindent \textbf{Qualitative Examples.} Fig.~\ref{fig:fig-teaser} showcases more complex cases beyond the primitive editing operations covered in the benchmark. DuoGen can execute intricate instructions that simultaneously modify backgrounds, adjust character appearance or clothing, change age or pose, and alter overall visual style. In addition, DuoGen supports combining multiple reference images with different resolutions, enabling flexible and compositional editing. Additional results are provided in the Supp.~\ref{supp-part:image-generation-editing-examples}.

\begin{table}[!t]
    \center
    \small
    \caption{\textbf{Comparison of different data strategies.} Abbreviation is aligned with Table~\ref{tab:comparison-comm-test}.}
    \addtolength{\tabcolsep}{-2.5pt}
    \begin{tabular}{l|cccccc}
    \toprule
    Data Configuration & Sty. & Enti. & Tren. & Comp. & ImgQ. & IRS  \\ \midrule
    CoMM original      & 6.14 & 6.21  & 6.52  & 6.45  & 6.30  & 4.42 \\
    w. Our data engine  & 7.85 & 7.76  & 7.22  & 8.15  & 7.79  & 5.91 \\
    + Synthetic Data   & 9.15 & 9.21  & 9.30  & 9.45  & 9.48  & 7.58 \\ \bottomrule
    \end{tabular}
    \label{tab:data-ablation}
\end{table}

\subsection{Data Ablation}
In this section, we evaluate the effectiveness of our data strategy using three configurations of instruction-tuning data on the CoMM benchmark~\cite{chen2025comm}: (1) the original CoMM data (with ~200k remaining samples due to expired image links); (2) CoMM data processed using our data engine; and (3) the processed CoMM data further augmented with our synthetic interleaved data. As shown in Table~\ref{tab:data-ablation}, applying our data engine yields substantial gains in both text quality and IRS (image-text alignment), highlighting the benefits of MLLM-based post-processing and cleaning. Incorporating synthetic data provides additional improvements, especially in image quality and temporal–semantic consistency.

\section{Conclusion}
We present DuoGen, a framework that advances interleaved multimodal generation through high-quality data, architecture design, training strategy, and quantitative benchmark. We curate 298k high-quality instruction-tuning samples from complementary sources for interleaved generation, along with large-scale interleaved context data for alignment. Instead of coupling uni-modal and interleaved generation during pretraining, DuoGen directly leverages a well-pretrained MLLM and a DiT pretrained on video generation, avoiding the uni-modal pretraining efforts. Finally, we introduce a dedicated benchmark for interleaved generation, enabling more comprehensive and standardized evaluation of this under-explored task.

\noindent \textbf{Limitations.} Our work primarily focuses on enabling and evaluating interleaved generation, leaving several directions for future exploration. These include scaling up the pretrained components for stronger reasoning and visual fidelity, as well as conducting deeper ablations on architectural choices—such as comparing cross-attention–based conditioning with MMDiT-style designs~\cite{sd3}.

{\small
\bibliographystyle{ieee_fullname}
\bibliography{egbib}
}

\onecolumn
\appendix

\section*{\centering Supplementary Materials}

\renewcommand{\numberline}[1]{#1.\enspace\space}
\startcontents[supp-part]
\printcontents[supp-part]{l}{1}{
}{}

\clearpage

\section{Interleaved Generation Examples}
\label{supp-part:interleaved-generation-examples}

\begin{figure}[h]
    \centering
    \includegraphics[width=0.95\linewidth]{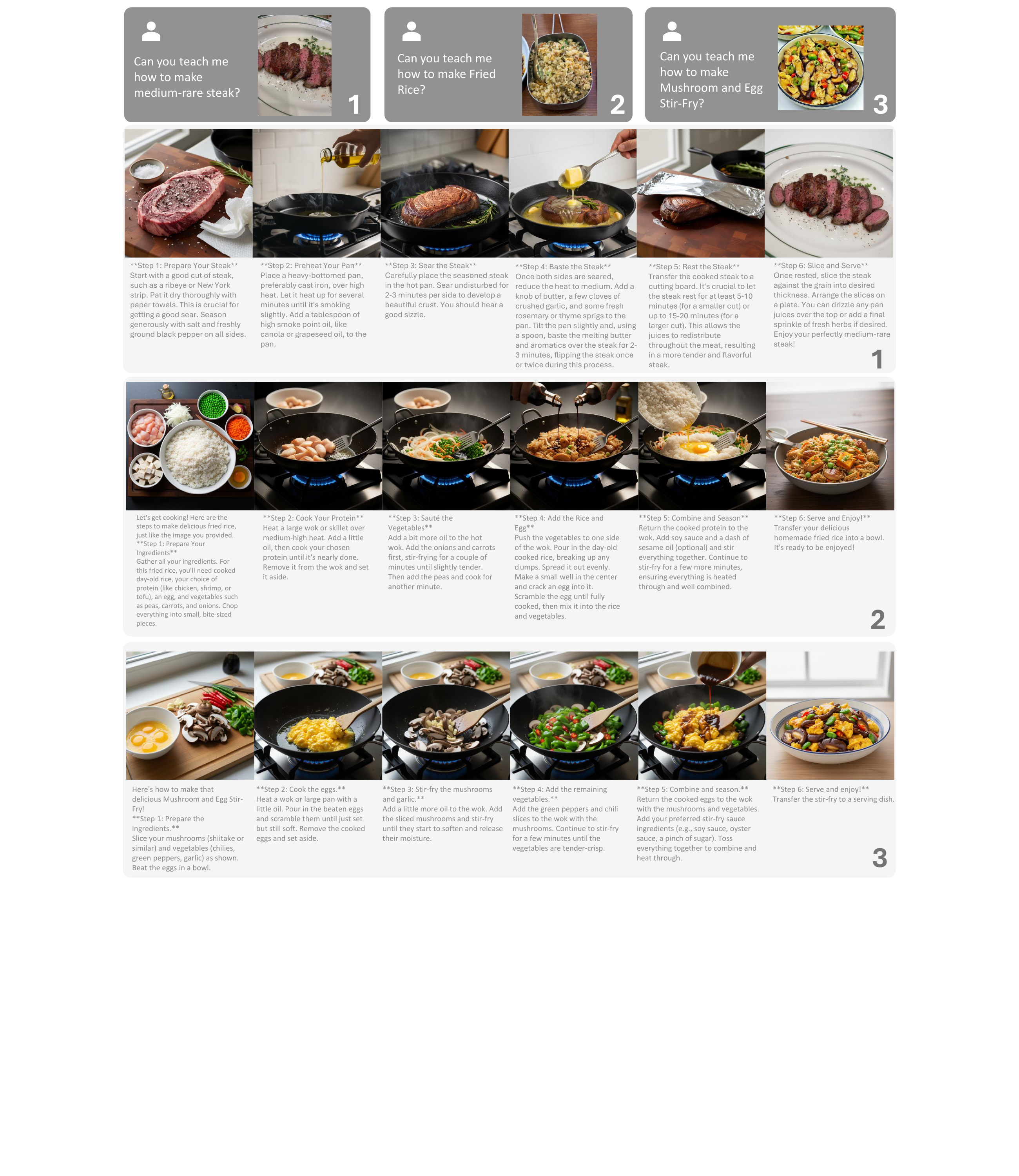}
     \caption{\textbf{Generation results of DuoGen on Cooking-200.}}  
    \label{supp-fig:interleaved-generation-cooking-part1}
\end{figure}

\begin{figure}[h]
    \centering
    \includegraphics[width=0.95\linewidth]{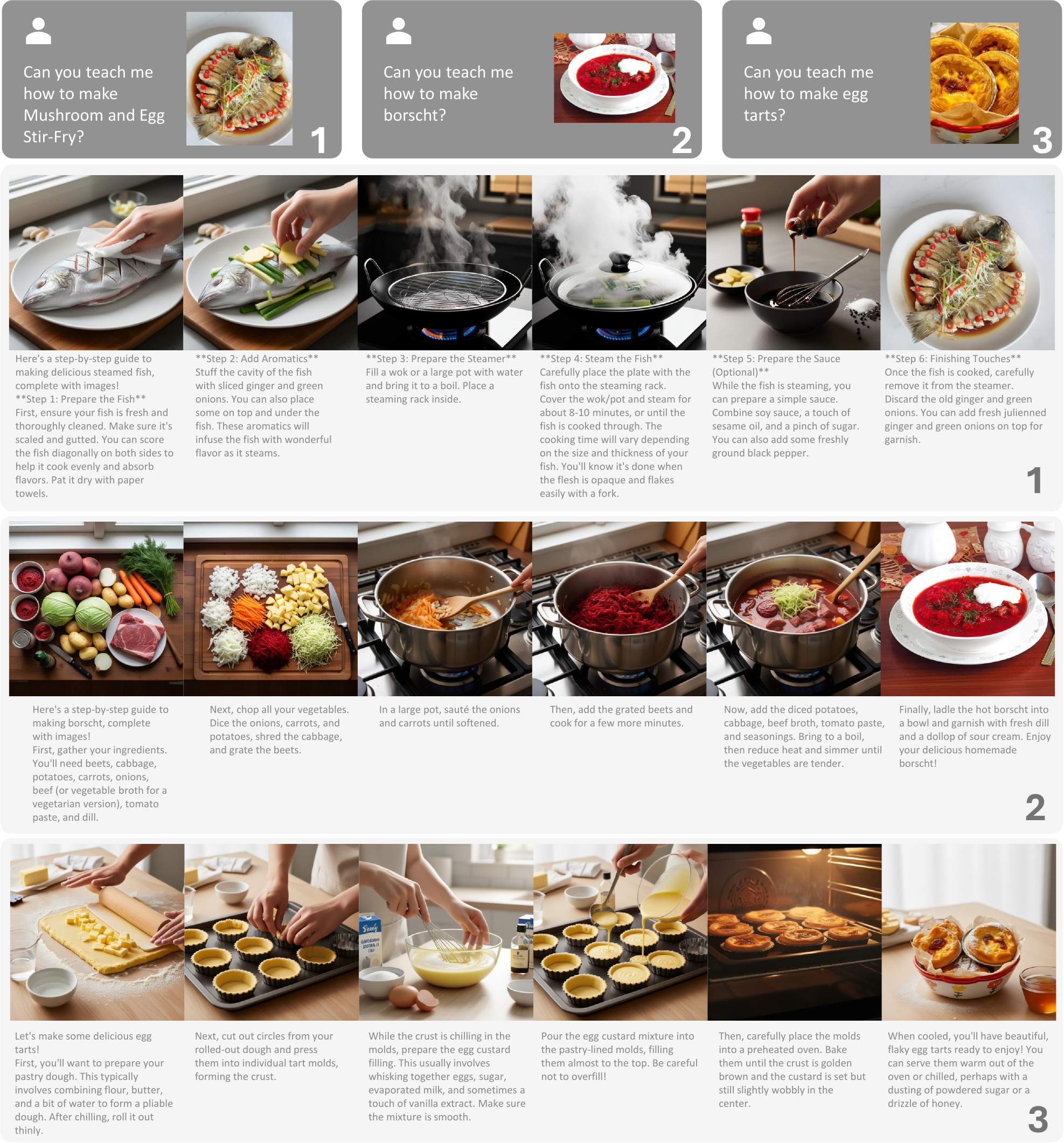}
     \caption{\textbf{Generation results of DuoGen on Cooking-200.}}  
    \label{supp-fig:interleaved-generation-cooking-part2}
\end{figure}

\begin{figure}[h]
    \centering
    \includegraphics[width=0.95\linewidth]{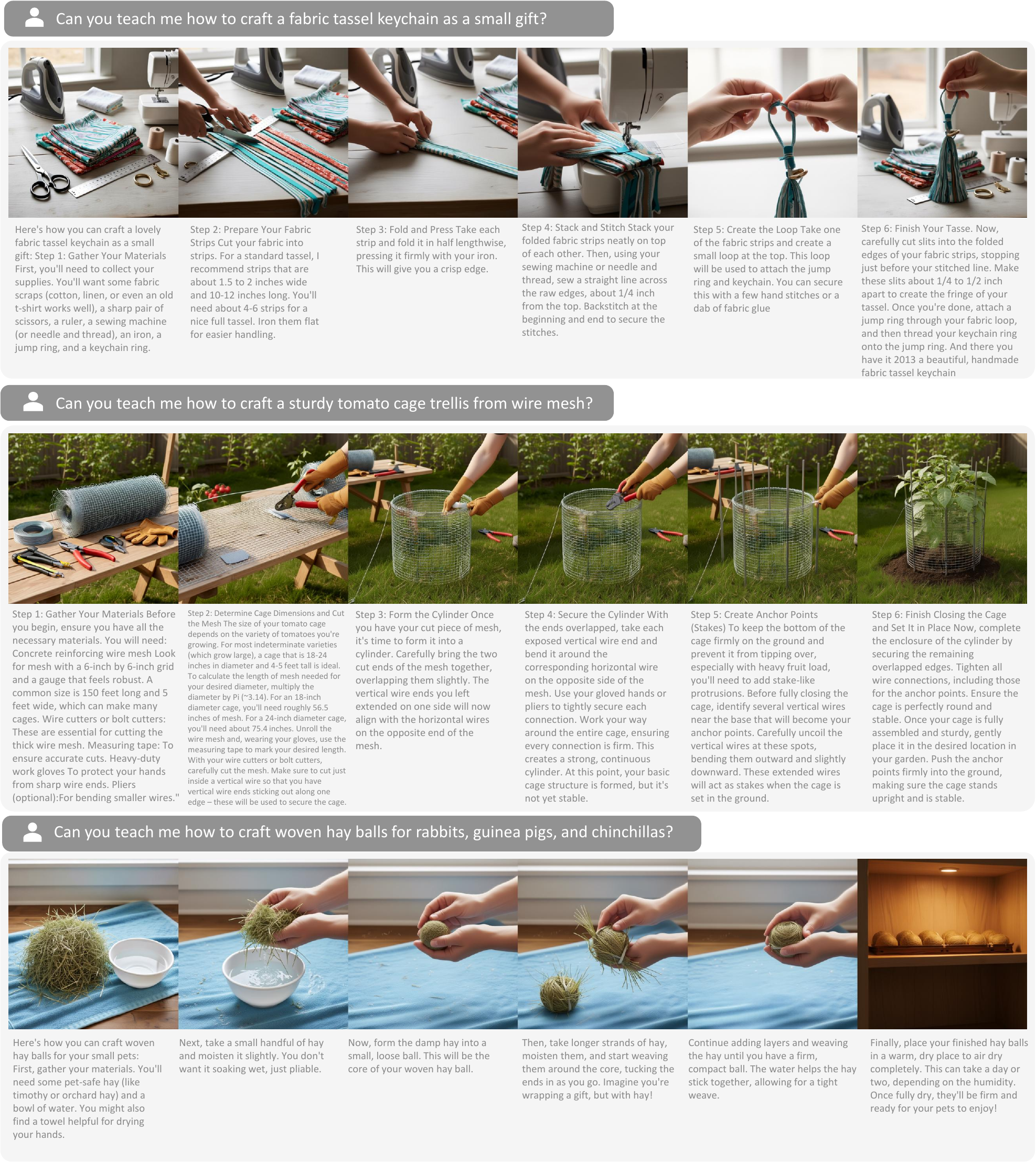}
     \caption{\textbf{Generation results of DuoGen on How-to-500.}}
    \label{supp-fig:interleaved-generation-general-part1}
\end{figure}

\begin{figure}[h]
    \centering
    \includegraphics[width=0.95\linewidth]{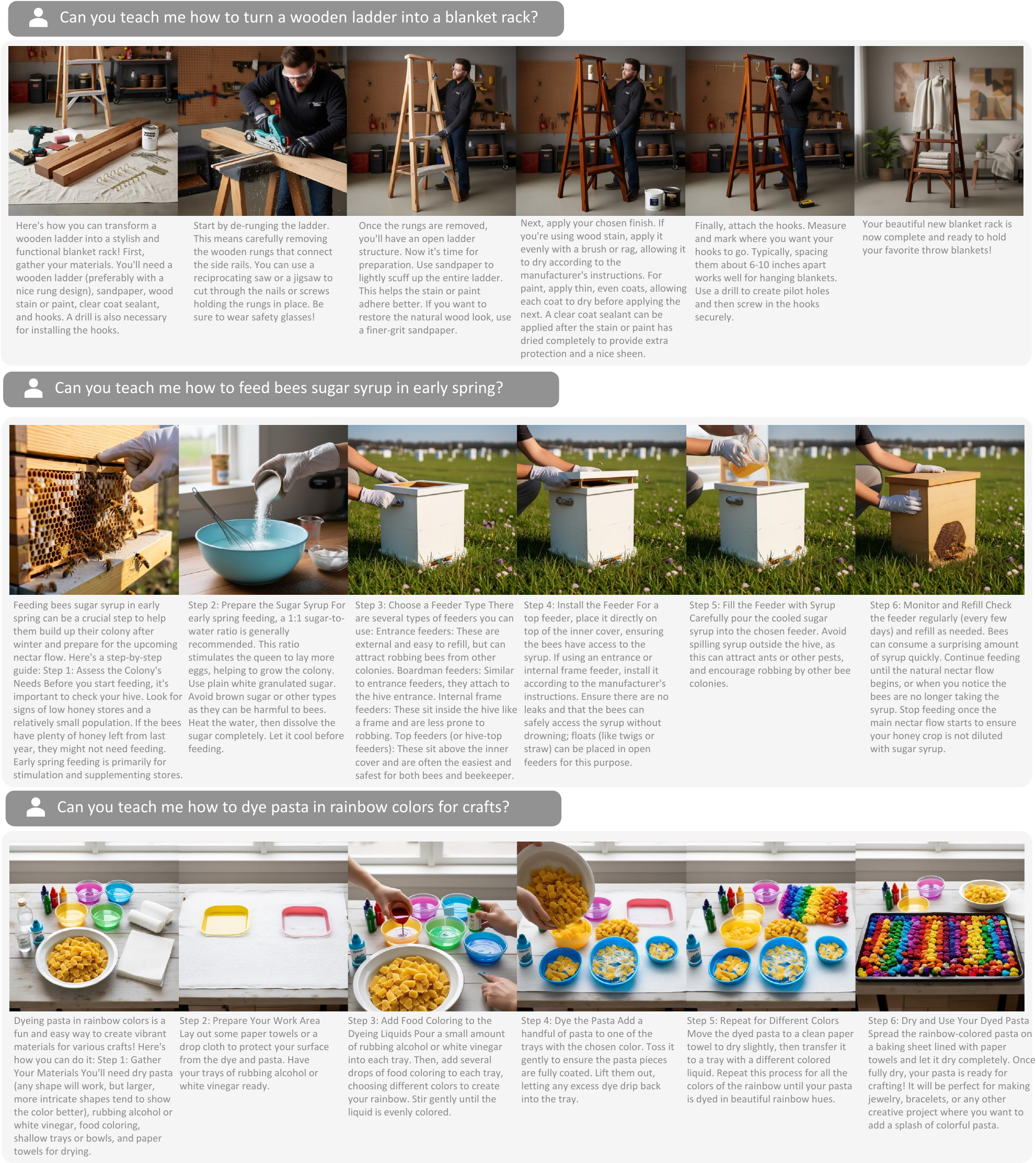}
     \caption{\textbf{Generation results of DuoGen on How-to-500.}}  
    \label{supp-fig:interleaved-generation-general-part2}
\end{figure}

\begin{figure}[h]
    \centering
    \includegraphics[width=0.95\linewidth]{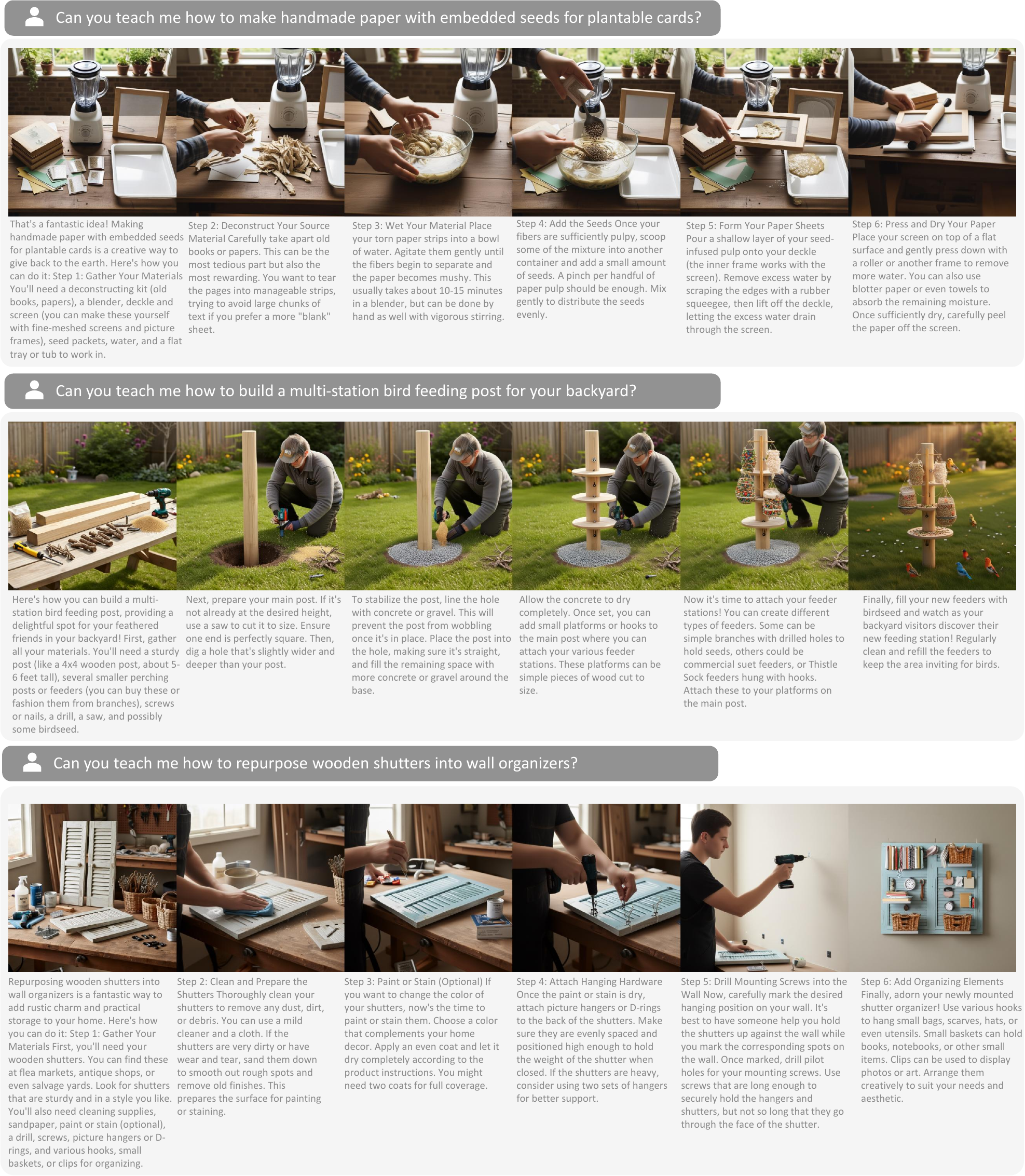}
     \caption{\textbf{Generation results of DuoGen on How-to-500.}}  
    \label{supp-fig:interleaved-generation-general-part3}
\end{figure}

\begin{figure}[h]
    \centering
    \includegraphics[width=0.95\linewidth]{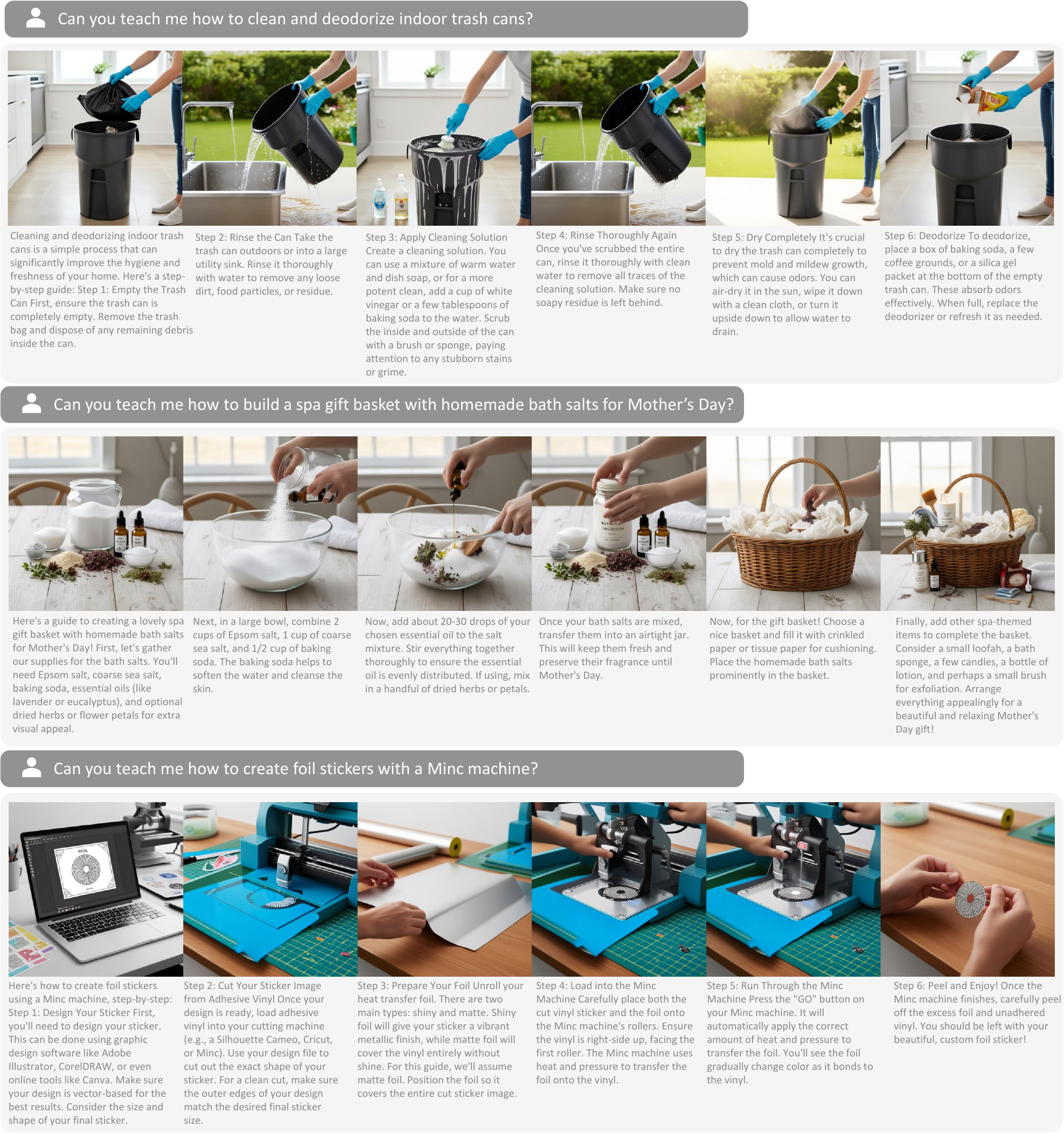}
     \caption{\textbf{Generation results of DuoGen on How-to-500.}}  
    \label{supp-fig:interleaved-generation-general-part4}
\end{figure}

\begin{figure}[h]
    \centering
    \includegraphics[width=0.95\linewidth]{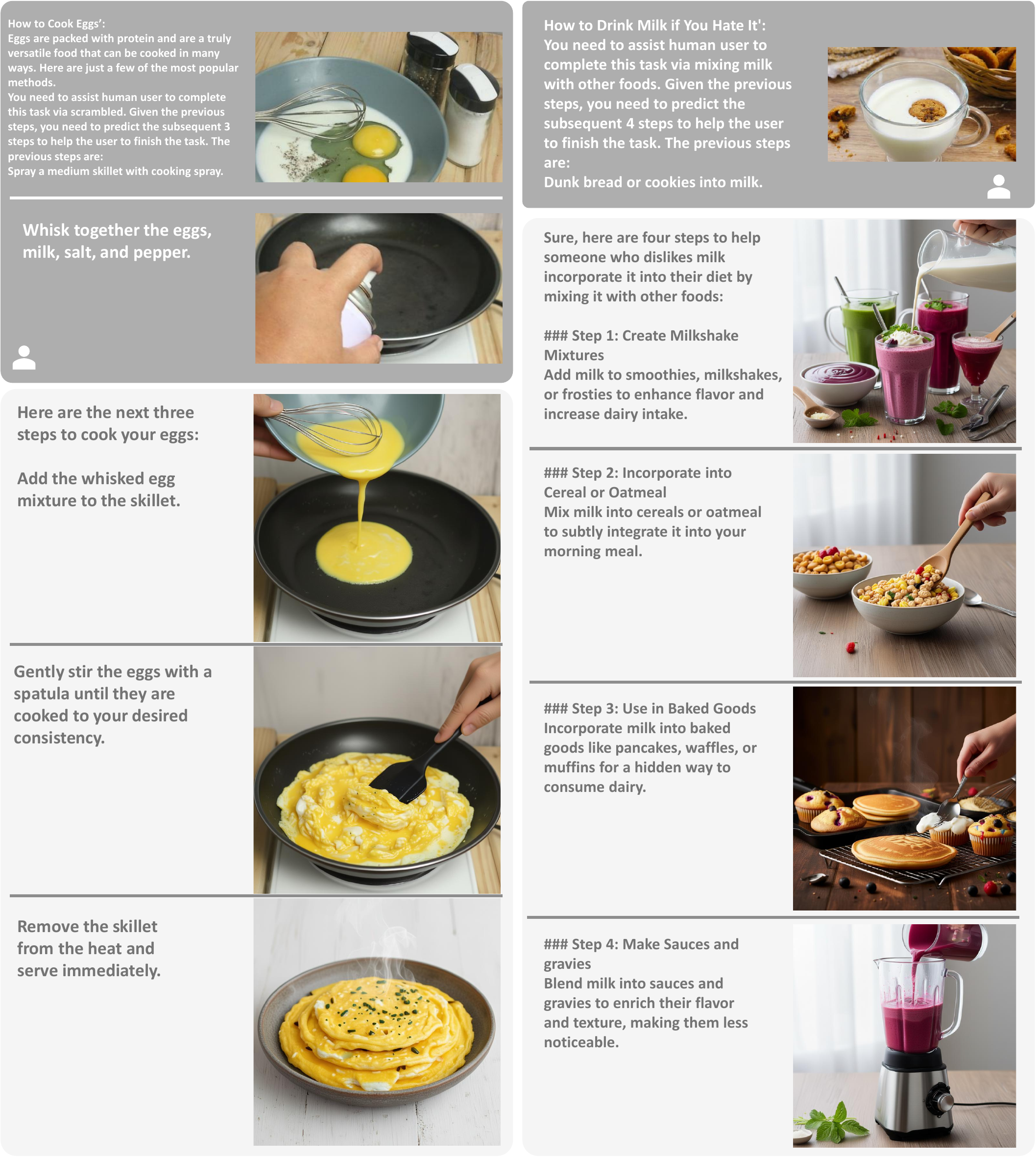}
     \caption{\textbf{Generation results of DuoGen on InterleavedBench.}}  
    \label{supp-fig:interleaved-generation-interleavedbench-part1}
\end{figure}

\begin{figure}[h]
    \centering
    \includegraphics[width=0.95\linewidth]{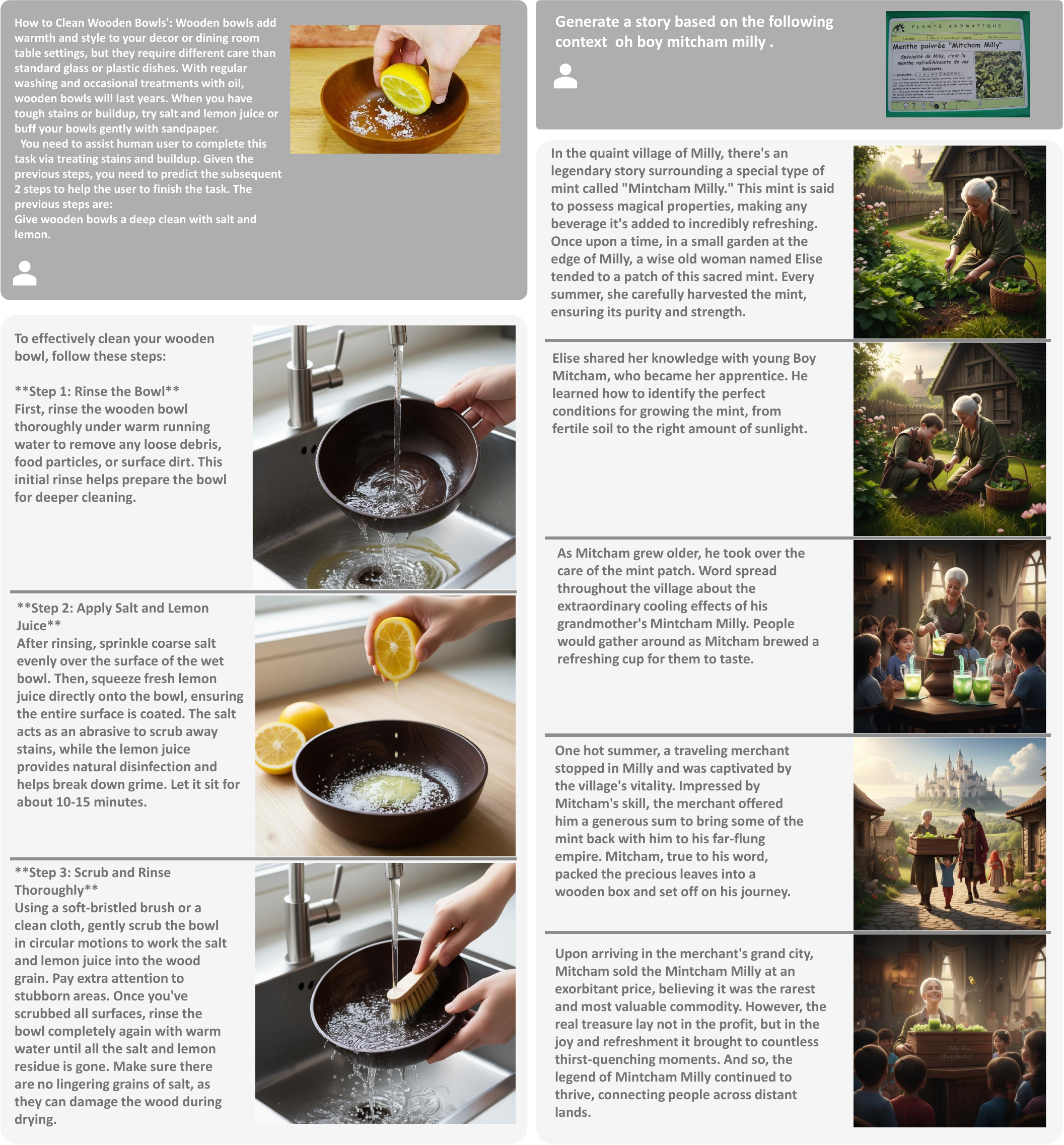}
     \caption{\textbf{Generation results of DuoGen on InterleavedBench.}}  
    \label{supp-fig:interleaved-generation-interleavedbench-part2}
\end{figure}

Fig.~\ref{supp-fig:interleaved-generation-cooking-part1} and Fig.~\ref{supp-fig:interleaved-generation-cooking-part2} present additional examples from the Cooking-200.
Fig.~\ref{supp-fig:interleaved-generation-general-part1}, Fig.~\ref{supp-fig:interleaved-generation-general-part2}, Fig.~\ref{supp-fig:interleaved-generation-general-part3}, and Fig.~\ref{supp-fig:interleaved-generation-general-part4} show examples from the How-to-500.
Fig.~\ref{supp-fig:interleaved-generation-interleavedbench-part1} and Fig.~\ref{supp-fig:interleaved-generation-interleavedbench-part2} provide results on InterleavedBench~\cite{interleavedbench}.

\clearpage

\section{Image Generation and Editing Examples}
\label{supp-part:image-generation-editing-examples}
\begin{figure}[h]
    \centering
    \includegraphics[width=0.95\linewidth]{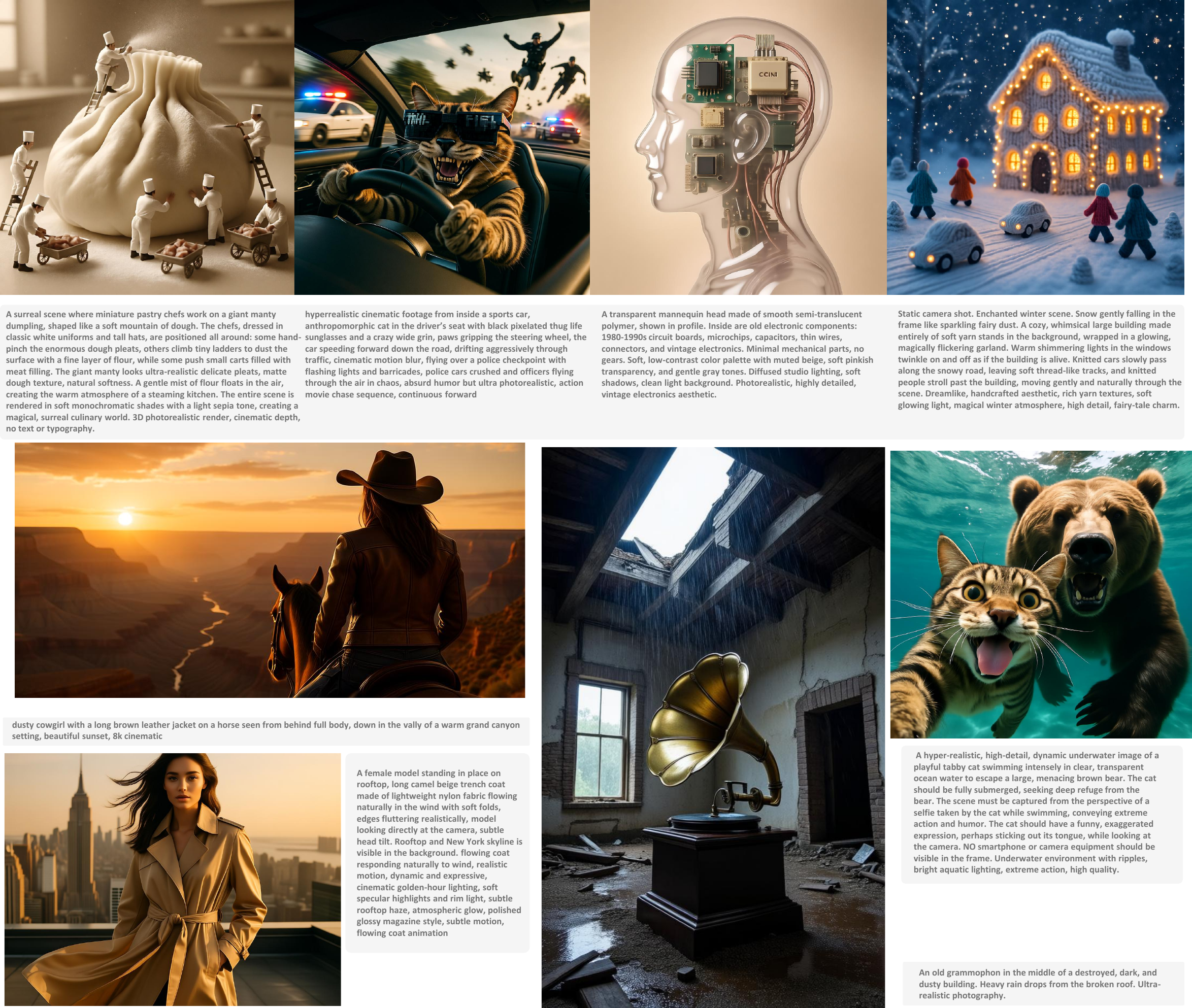}
     \caption{\textbf{Text-to-Image generation results of DuoGen.}}  
    \label{supp-fig:text-to-image-examples}
\end{figure}

\begin{figure}[h]
    \centering
    \includegraphics[width=0.95\linewidth]{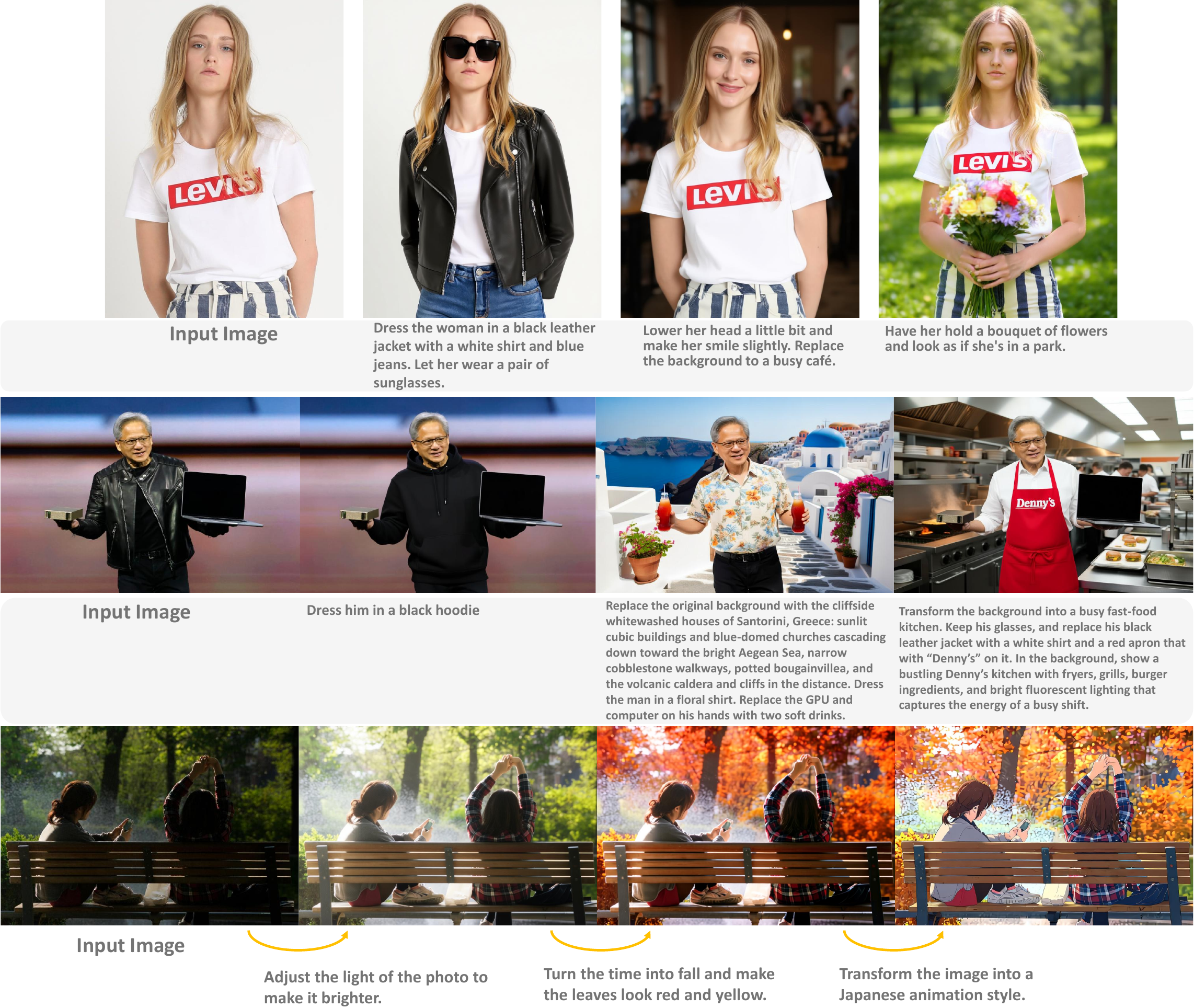}
     \caption{\textbf{Image editing results of DuoGen.}}  
    \label{supp-fig:image-editing-examples}
\end{figure}

\begin{figure}[h]
    \centering
    \includegraphics[width=0.95\linewidth]{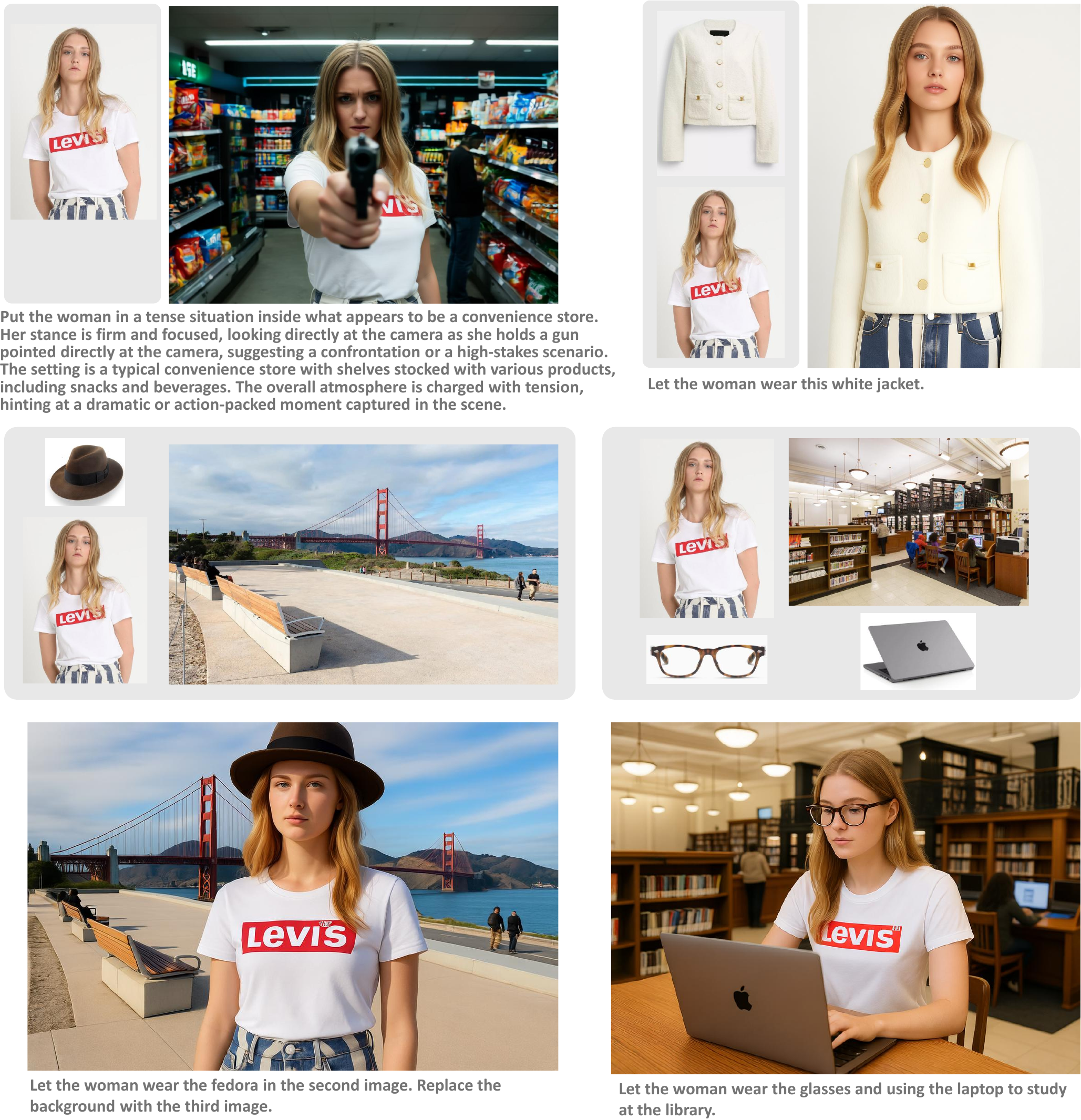}
     \caption{\textbf{Multi-reference image generation results of DuoGen.}}  
    \label{supp-fig:multi-reference-generation-examples}
\end{figure}

Fig.~\ref{supp-fig:text-to-image-examples}, Fig.~\ref{supp-fig:image-editing-examples}, and Fig.~\ref{supp-fig:multi-reference-generation-examples} shows more examples on image generation, image editing, and multi-reference image generation, respectively.

\clearpage

\section{Detailed Results on Image Generation and Editing Benchmarks}
\label{supp-part:detailed-quantitative-results-generation}

Due to space limitations in the main paper, we reported only the overall scores for the GenEval~\cite{geneval} and ImgEdit~\cite{ye2025imgedit} benchmarks. Here, we provide the detailed scores for each subtask.
Table~\ref{supp-tab:comparison-t2i-geneval} presents the complete GenEval results, and Table~\ref{supp-tab:comparison-editing-imgedit} provides the detailed results for ImgEdit.

\begin{table*}[h]
    \center
    \small
    \addtolength{\tabcolsep}{-2pt}
    \begin{tabular}{c|lccccccc}
    \toprule
    Model Type                                                                                       & Method        & Single Object & Two Object & Counting & Colors & Position & Attribute Binding & Overall \\ \midrule
    Commercial                                                                                       & GPT-4o-Image~\cite{openai-gpt4o-2024}  & 0.99          & 0.92       & 0.85     & 0.92   & 0.75     & 0.61              & 0.84    \\ \midrule
    \multirow{4}{*}{Generation}                                                                      & SDXL~\cite{podell2023sdxl}          & 0.98          & 0.74       & 0.39     & 0.85   & 0.15     & 0.23              & 0.55    \\
                                                                                                     & DALLE-3~\cite{openai_dalle3_2023}       & 0.96          & 0.87       & 0.47     & 0.83   & 0.43     & 0.45              & 0.67    \\
                                                                                                     & FLUX.1-dev~\cite{flux2024}    & 0.98          & 0.93       & 0.75     & 0.93   & 0.68     & 0.65              & 0.82    \\
                                                                                                     & Qwen-Image~\cite{wu2025qwen}    & 0.99          & 0.92       & 0.89     & 0.88   & 0.76     & 0.77              & 0.87    \\ \midrule
    \multirow{9}{*}{Unified Model}                                                                   & Emu3~\cite{wang2024emu3}          & 0.98          & 0.71       & 0.34     & 0.81   & 0.17     & 0.21              & 0.54    \\
                                                                                                     & Show-o~\cite{showo}        & 0.95          & 0.52       & 0.49     & 0.82   & 0.11     & 0.28              & 0.53    \\
                                                                                                     & Janus-Pro-7B~\cite{chen2025janus}  & 0.99          & 0.89       & 0.59     & 0.9    & 0.79     & 0.66              & 0.8     \\
                                                                                                     & MMaDA~\cite{yang2025mmada}         & 0.99          & 0.76       & 0.61     & 0.84   & 0.2      & 0.37              & 0.63    \\
                                                                                                     & MetaQuery-XL*~\cite{pan2025transfer} & -             & -          & -        & -      & -        & -                 & 0.8     \\
                                                                                                     & BLIP3o~\cite{chen2025blip3}       & -             & -          & -        & -      & -        & -                 & 0.84    \\
                                                                                                     & Bagel~\cite{deng2025emerging}         & 0.99          & 0.94       & 0.81     & 0.88   & 0.64     & 0.63              & 0.82    \\
                                                                                                     & UniWorld-V1~\cite{lin2025uniworld}   & 0.99          & 0.93       & 0.79     & 0.89   & 0.49     & 0.7               & 0.8     \\
                                                                                                     & OmniGen2~\cite{wu2025omnigen2}      & 1             & 0.95       & 0.64     & 0.88   & 0.55     & 0.76              & 0.8     \\ \midrule
    \multirow{2}{*}{\begin{tabular}[c]{@{}c@{}}Unified Interleaved \\ Generation Model\end{tabular}} & Uni-CoT**~\cite{qin2025unicot}     & 0.99          & 0.96       & 0.84     & 0.92   & 0.57     & 0.71              & 0.83    \\
                                                                                                     & Ours          &   0.82    & 0.99   &  0.94   & 0.91   & 0.84    &  0.80   &  0.88         \\ \bottomrule
    \end{tabular}
    \caption{\textbf{Detailed comparison on GenEval benchmark.} * denotes that the model is using LLM to rewrite prompts. ** Uni-CoT is using interleaved generation to improve image generation quality.}
    \label{supp-tab:comparison-t2i-geneval}
\end{table*}

\begin{table*}[h]
    \center
    \small
    \addtolength{\tabcolsep}{-3pt}
    \begin{tabular}{c|lcccccccccc}
    \toprule
    Model Type                     &                          & Add  & Adjust & Extract & Replace & Remove & Background & Style & Hybrid & Action & Overall \\ \midrule
    Commercial Model               & GPT-4o-Image~\cite{openai-gpt4o-2024}             & 4.61 & 4.33   & 2.90    & 4.35    & 3.66   & 4.57       & 4.93  & 3.96   & 4.89   & 4.20    \\ \midrule
    \multirow{4}{*}{Generation}    & ICEdit~\cite{zhang2025context}                   & 3.58 & 3.39   & 1.73    & 3.15    & 2.93   & 3.08       & 3.84  & 2.04   & 3.68   & 3.05    \\
                                   & Step1X-Edit~\cite{liu2025step1x}              & 3.88 & 3.14   & 1.76    & 3.40    & 2.41   & 3.16       & 4.63  & 2.64   & 2.52   & 3.06    \\
                                   & FLUX.1 Kontext {[}Pro{]}~\cite{labs2025flux} & 4.25 & 4.15   & 2.35    & 4.56    & 3.57   & 4.26       & 4.57  & 3.68   & 4.63   & 4.00    \\
                                   & Qwen-Image~\cite{wu2025qwen}               & 4.38 & 4.16   & 3.43    & 4.66    & 4.14   & 4.38       & 4.81  & 3.82   & 4.69   & 4.27    \\ \midrule
    \multirow{6}{*}{Unified Model} & OmniGen~\cite{omnigen}                  & 3.47 & 3.04   & 1.71    & 2.94    & 2.43   & 3.21       & 4.19  & 2.24   & 3.38   & 2.96    \\
                                   & Bagel~\cite{deng2025emerging}                    & 3.56 & 3.31   & 1.70    & 3.30    & 2.62   & 3.24       & 4.49  & 2.38   & 4.17   & 3.20    \\
                                   & UniWord-V1~\cite{lin2025uniworld}               & 3.82 & 3.64   & 2.27    & 3.47    & 3.24   & 2.99       & 4.21  & 2.96   & 2.74   & 3.26    \\
                                   & OmniGen2~\cite{wu2025omnigen2}                 & 3.57 & 3.06   & 1.77    & 3.74    & 3.20   & 3.57       & 4.81  & 2.52   & 4.68   & 3.44    \\
                                   & OVIS-U1~\cite{wang2025ovis}                  & 4.13 & 3.62   & 2.98    & 4.45    & 4.06   & 4.22       & 4.69  & 3.45   & 4.61   & 4.00    \\
                                   & Lumina-DiMOO~\cite{xin2025lumina}             & 3.82 & -      & -       & 3.83    & 2.76   & -          & 4.18  & -      & -      & -       \\ \midrule
    Interleaved         & DuoGen &      4.53&        4.33&         2.28&         4.69&        4.71&            4.61&       4.51&        3.85&        4.67&         4.19\\ \bottomrule
    \end{tabular}
    \caption{\textbf{Detailed comparison on ImgEdit benchmark.}}
    \label{supp-tab:comparison-editing-imgedit}
\end{table*}

\clearpage

\section{Details of the Interleaved Generation Benchmark}
\label{supp-part:interleaved-benchmark-details}

Below is the prompt used to evaluate model outputs on our interleaved generation benchmark.
\begin{figure}[h]
  \centering
  \includegraphics[width=0.90\linewidth]{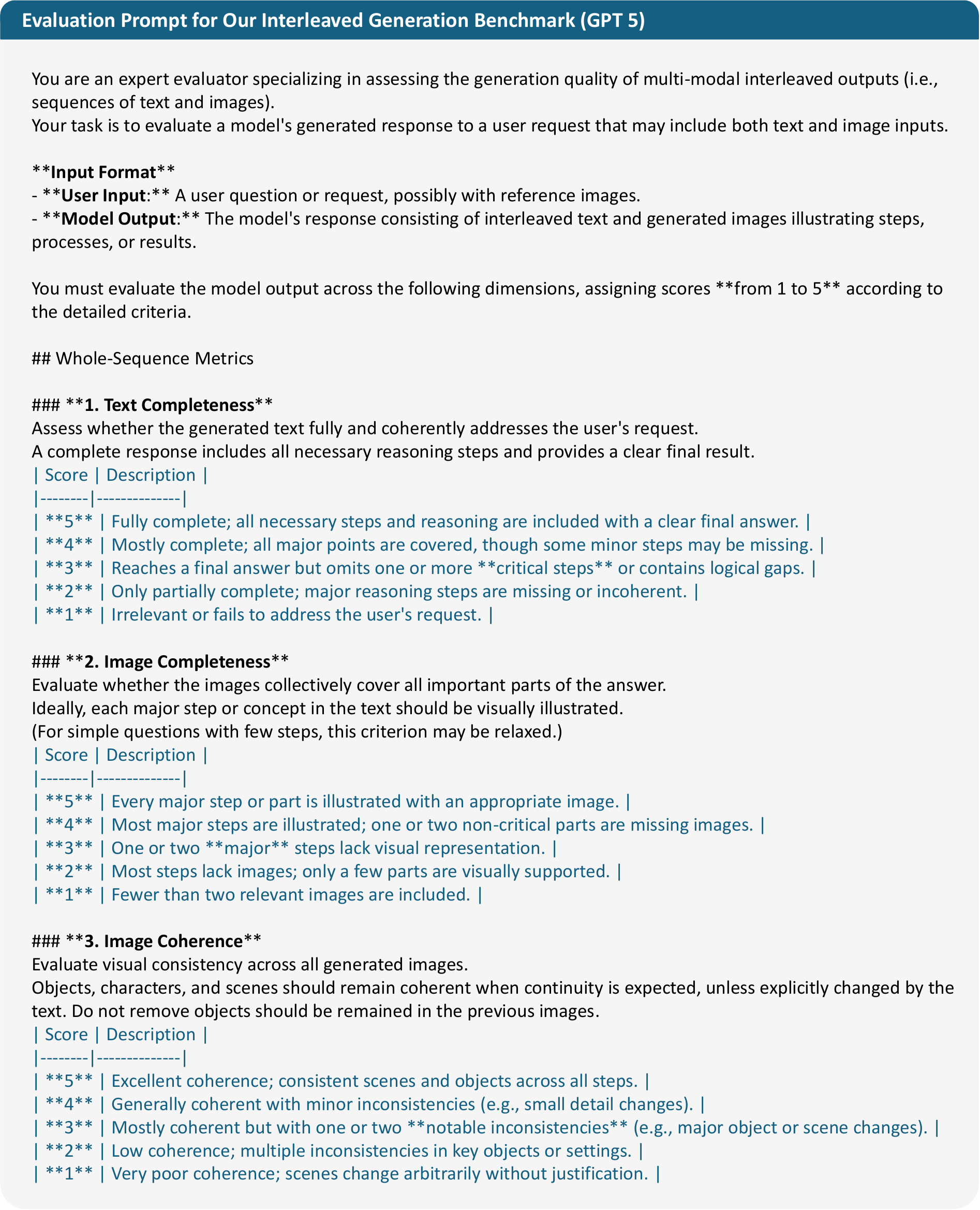}
\end{figure}
\begin{figure}[h]
  \centering
  \includegraphics[width=0.88\linewidth]{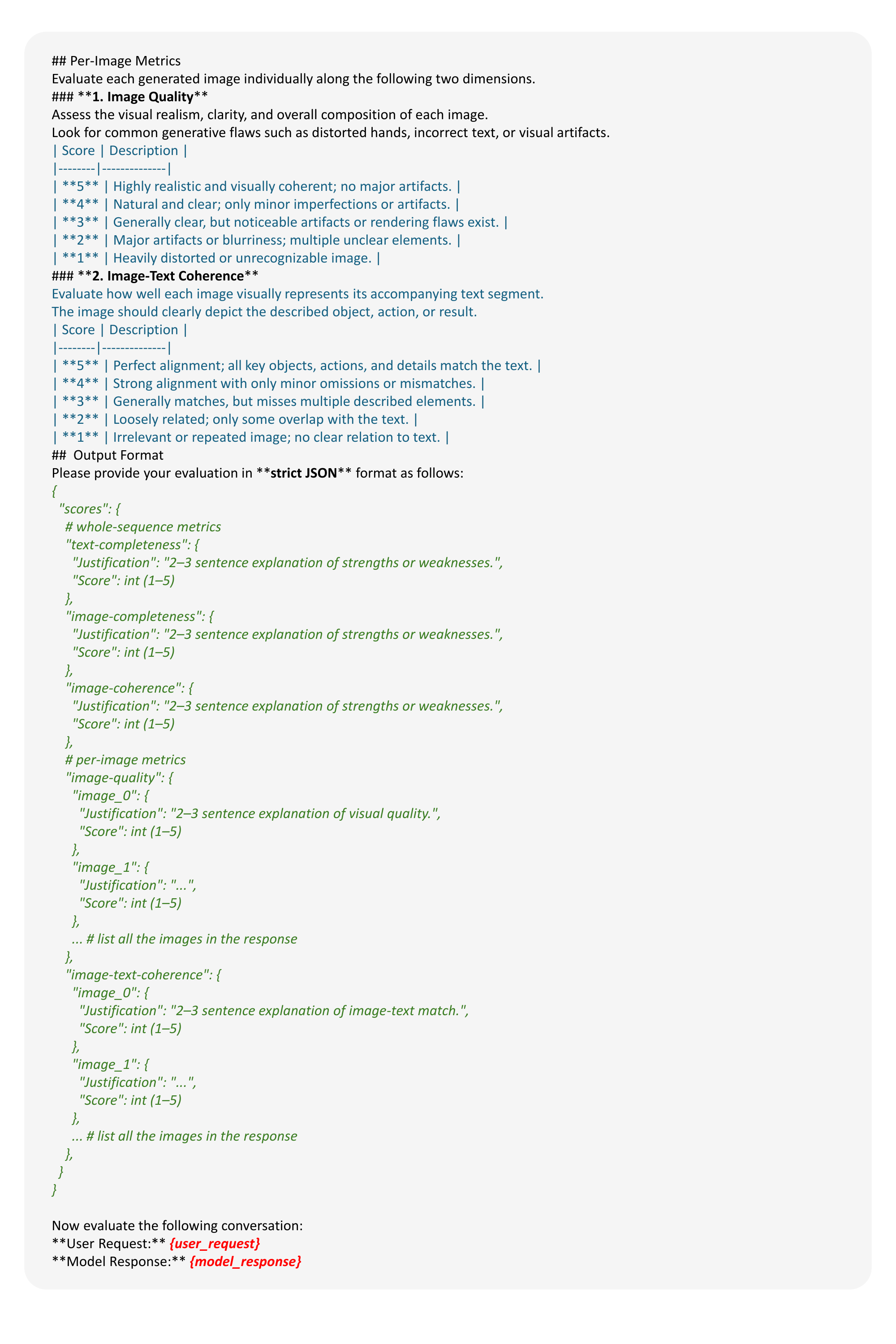}
\end{figure}

\clearpage

\section{Data Engine for Webpages}
\label{supp-part:webpage-data-engine}

As discussed in Sec.~3.1 of the main paper, we convert raw web pages into multi-modal conversations using a data engine with a series of cleaning and rewriting steps. The full pipeline consists of four major stages: 1) rewriting and splitting the raw interleaved content; 2) captioning and categorizing all images; 3) removing duplicate images and reordering the interleaved text–image sequence; and 4) converting the cleaned interleaved content into a user–assistant conversation.

Below, we present the prompt used in the first stage, which transforms the raw webpage into introduction and main content, both represented as interleaved image–text sequences. In this step, the VLM also rewrites the content to correct errors and remove irrelevant elements such as advertisements. To improve throughput, we replace image content with text placeholders, as this step focuses solely on text rewriting rather than image understanding.

\begin{figure}[h]
  \centering
  \includegraphics[width=0.99\linewidth]{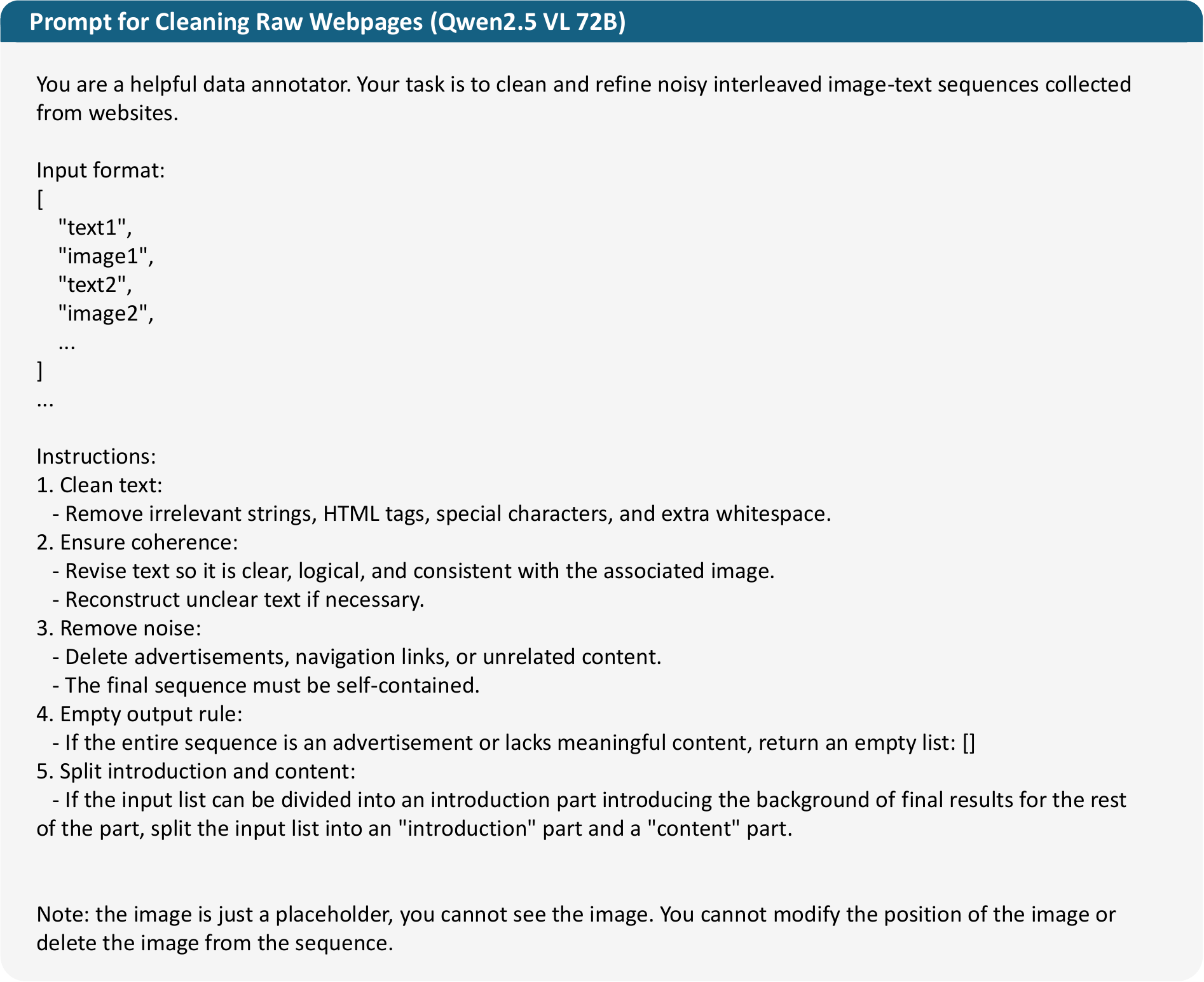}
\end{figure}
\clearpage

\vspace{-10pt}
\begin{figure}[h]
  \centering
  \includegraphics[width=0.90\linewidth]{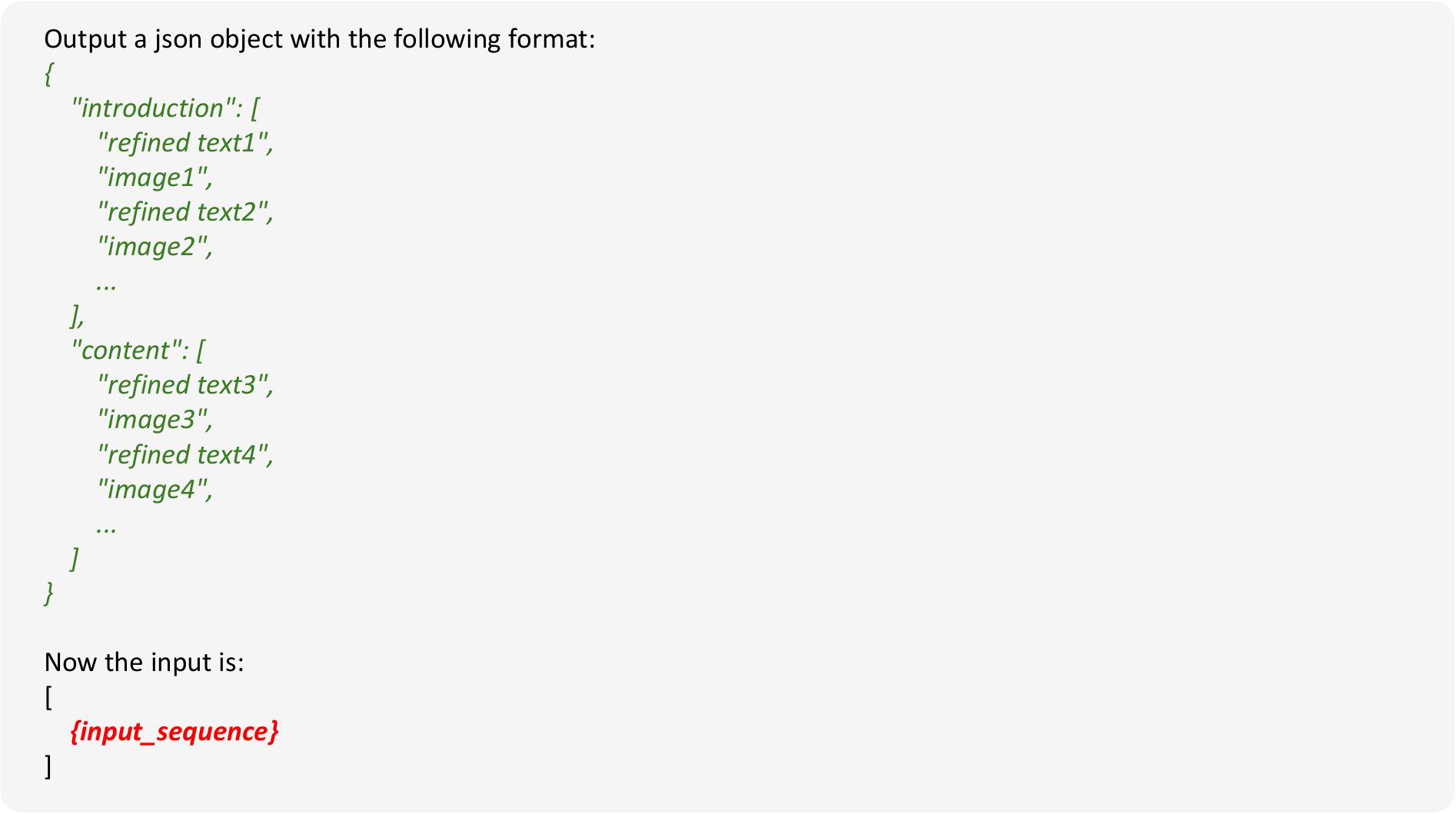}
\end{figure}
\vspace{-5pt}

Below is the prompt used to caption and classify the images in the interleaved sequence.
\begin{figure}[h]
  \centering
  \includegraphics[width=0.90\linewidth]{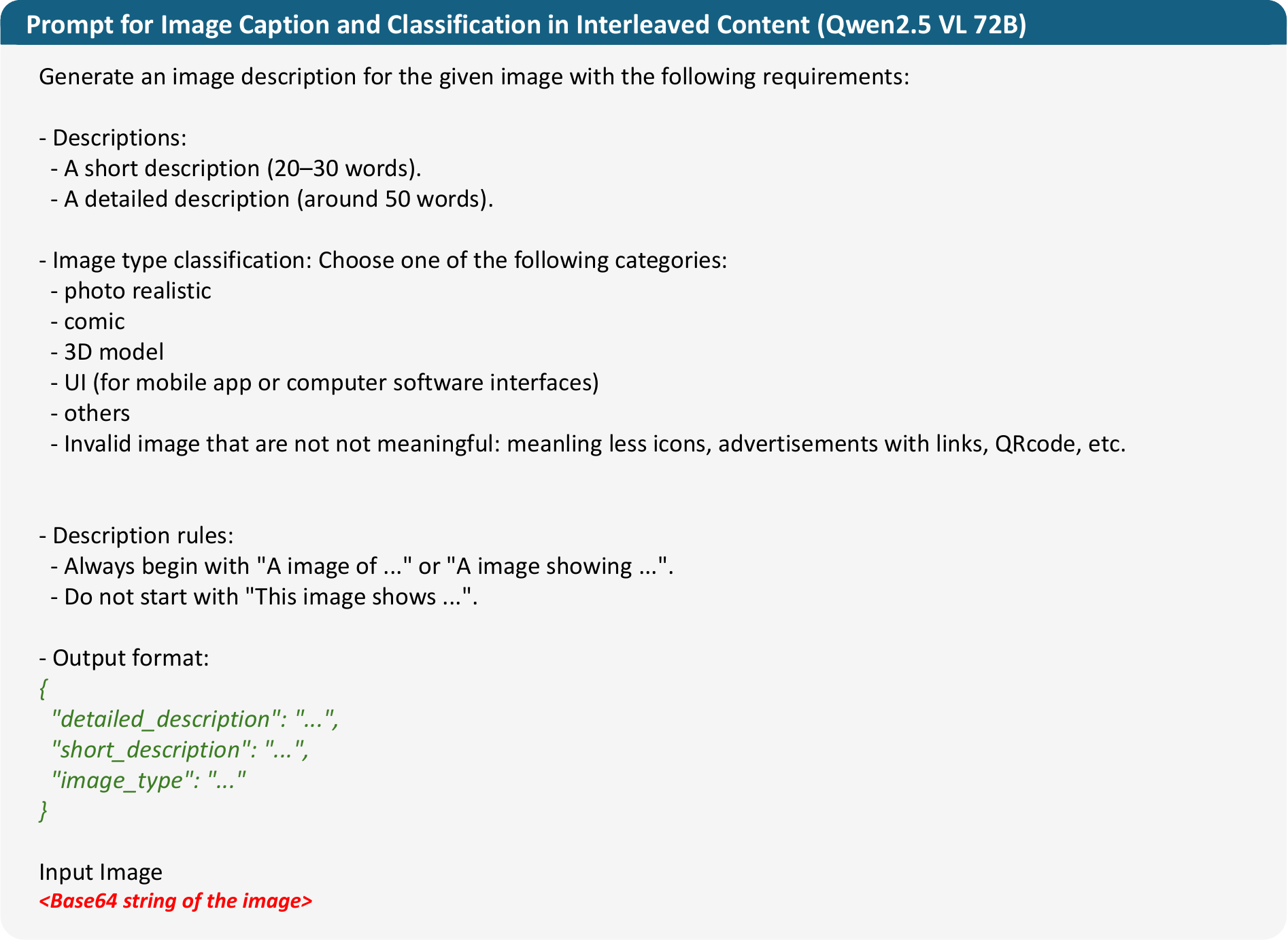}
\end{figure}
\vspace{-10pt}

Below is the prompt used to remove duplicate images and to reorder the interleaved image and text chunks.
\begin{figure}[h]
  \centering
  \includegraphics[width=0.95\linewidth]{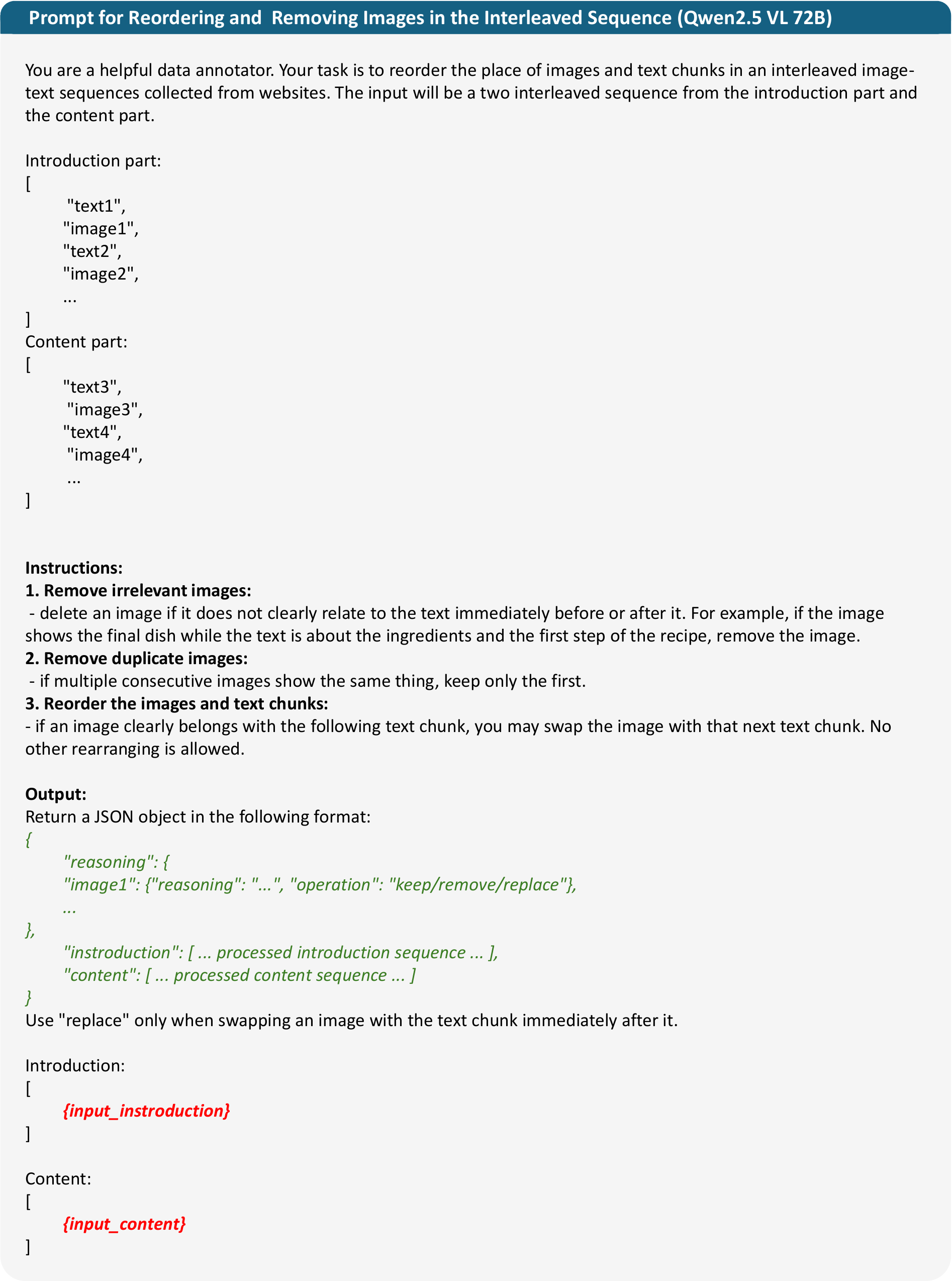}
\end{figure}
\clearpage

After the steps described above, we obtain a cleaned interleaved sequence. We then prompt the VLM with the following instruction to convert this sequence into a user–assistant conversation.

\begin{figure}[h]
  \centering
  \includegraphics[width=0.95\linewidth]{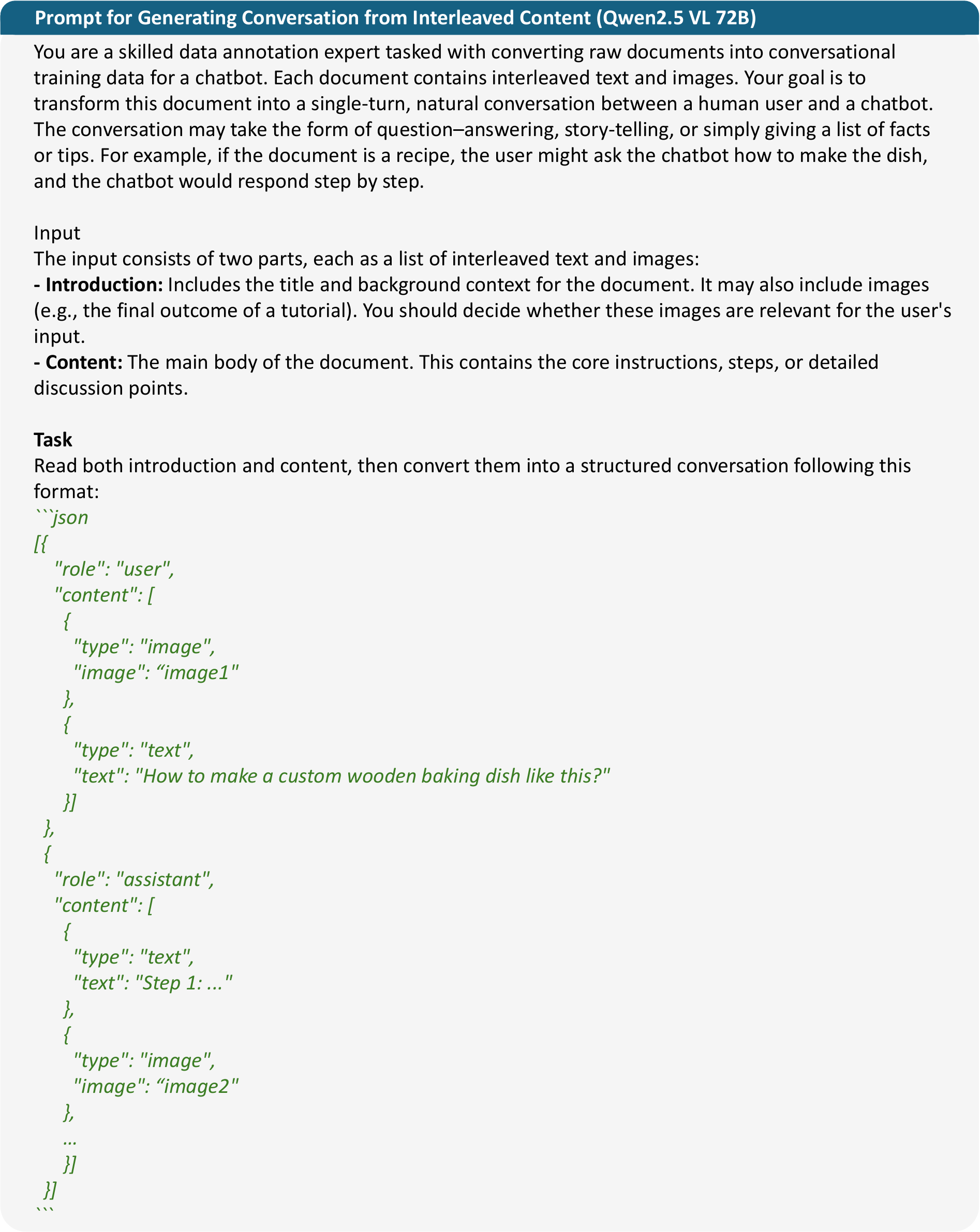}
\end{figure}

\begin{figure}[!t]
  \centering
  \includegraphics[width=0.95\linewidth]{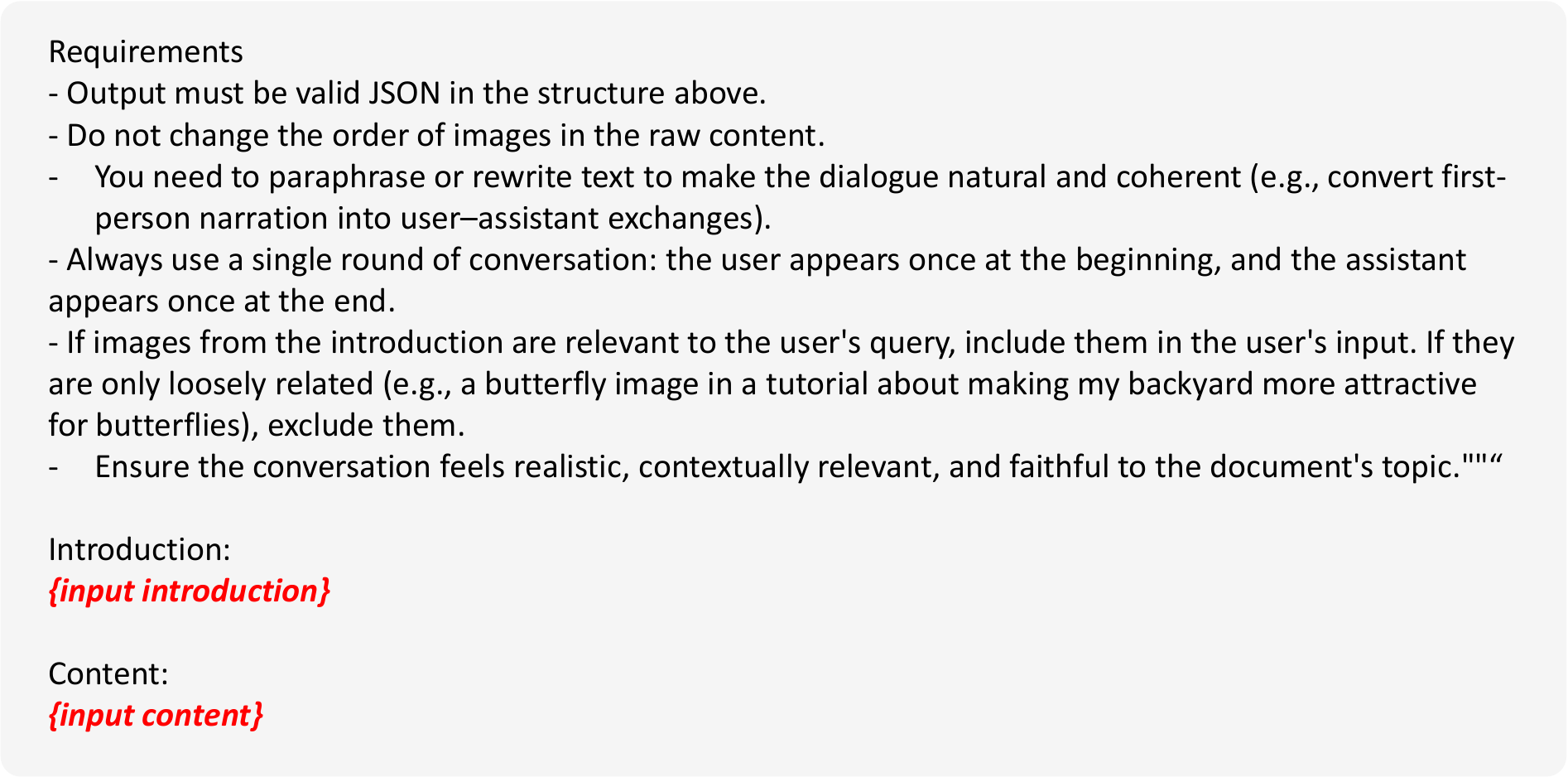}
\end{figure}

Note that in this step, only the images from the introduction section are fed to the VLM, allowing it to determine whether and where images should appear in the user's query. All images in the main content remain as placeholders when provided to the VLM.

\clearpage

\section{Synthetic Interleaved Generation Data}
\label{supp-part:synthetic-data-prompt}

To construct the prompts for general-purpose instruction-tuning data, we start from eight base categories: ``\textit{Sports}'', ``\textit{Outdoor \& Survival}'', ``\textit{DIY \& Crafting}'', ``\textit{Vehicle \& Transportation}'', ``\textit{Personal Care \& Health}'', ``\textit{Farm, Pet, and Animals}'', ``\textit{Home \& Living}'', and ``\textit{Office \& Productivity}''. We ask eight human annotators to further decompose these base categories into 151 subcategories and to write approximately 10 seed questions for each subcategory.

We then prompt OpenAI O3~\cite{o3-system-card-2025} with the instruction shown below to expand these seed questions, yielding 15{,}270 questions in total. To reduce duplication and improve generation quality, we use the highest thinking budget and provide the model with the other subcategories and their seed questions within the same base category as additional context.

\begin{figure}[h]
  \centering
  \includegraphics[width=0.90\linewidth]{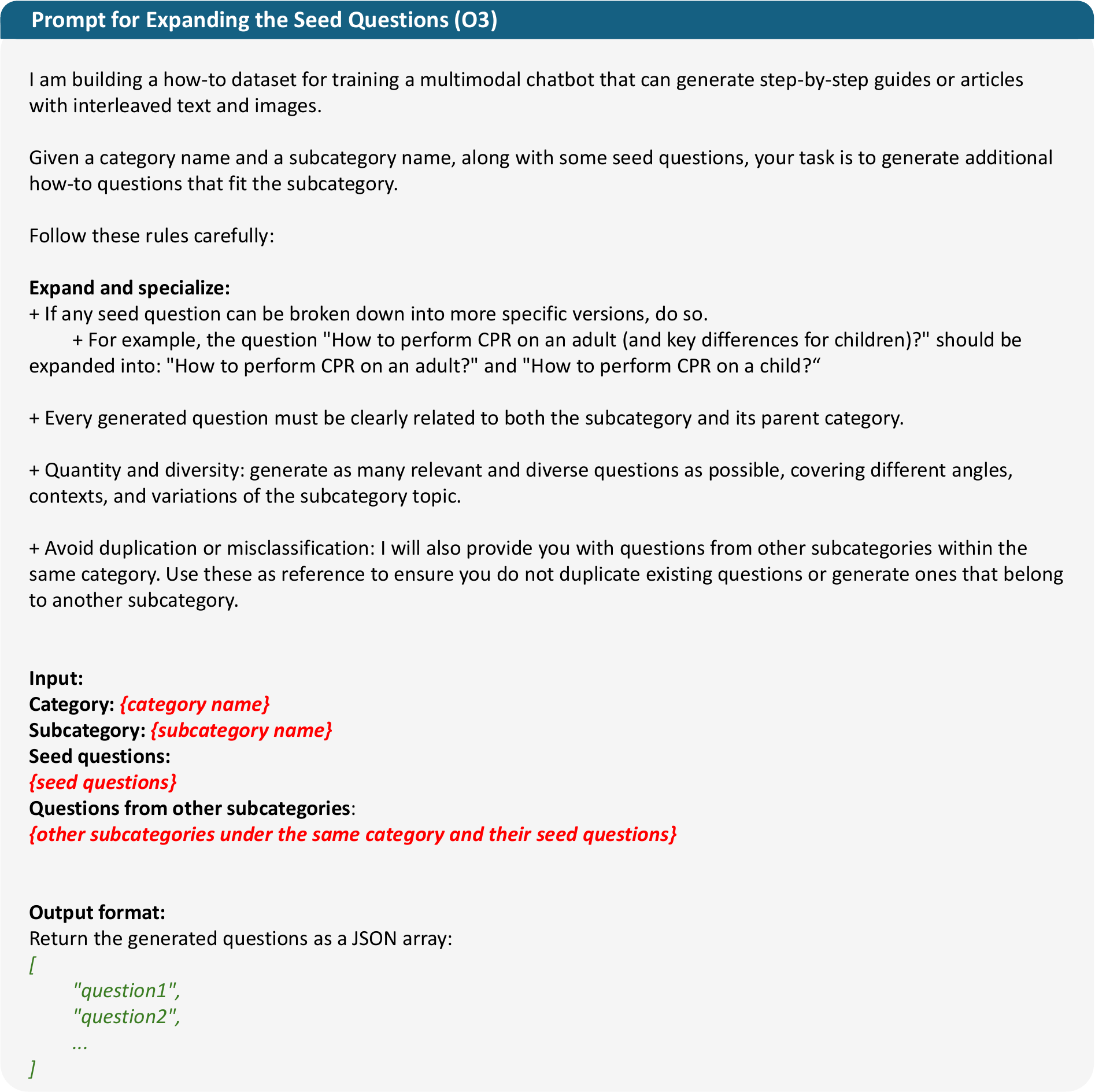}
\end{figure}

\clearpage

\begin{figure}[!t]
    \centering
    \includegraphics[width=0.75\linewidth]{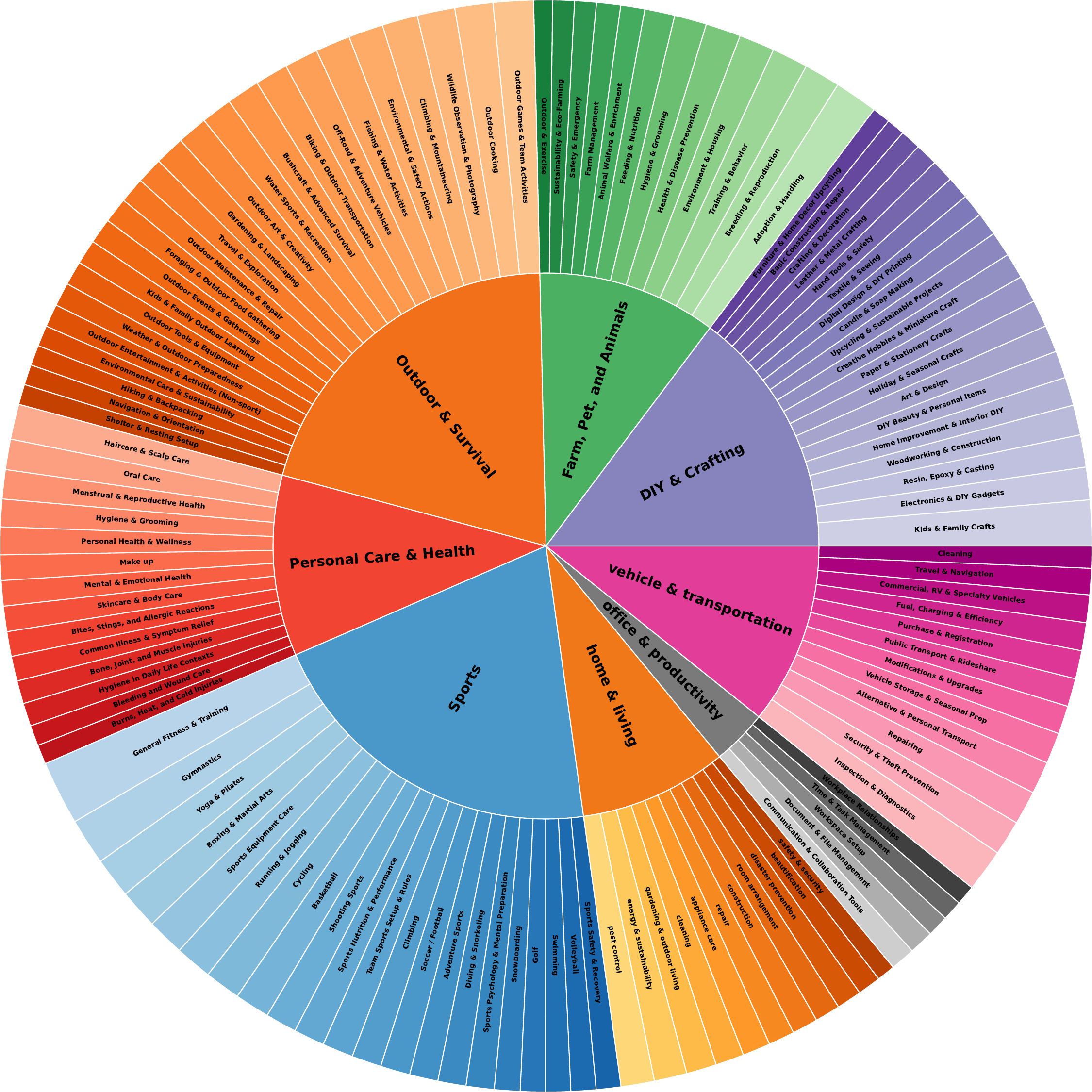}
     \caption{\textbf{All the basic and subcategories of the synthetic interleaved generation data and the number of prompts under it.}}  
    \label{supp-fig:synthetic-data-prompt-categories}
\end{figure}
The full set of subcategories, along with the proportional distribution of their prompts, is visualized in the pie chart in Fig.~\ref{supp-fig:synthetic-data-prompt-categories}.


For cooking-related data, we adopt MM-Food-100k~\cite{mmfood}. We filter out entries with invalid images or invalid dish names (e.g., raw ingredients or multiple dishes), randomly sample 15k valid examples, and then recaption the dish names using Qwen2.5 VL 72B~\cite{bai2025qwen2}.

\clearpage

\section{Interleaved Context Alignment Data}
\label{supp-part:interleaved-context-alignment-data}

Table~\ref{supp-tab:interleaved-context-alignment-data} lists all datasets used in the interleaved context-alignment stage. We adopt open-source datasets spanning text-to-image generation, image editing, and multi-reference generation tasks.

\begin{table*}[h]
    \center
    \small
    \addtolength{\tabcolsep}{-2pt}
    \begin{tabular}{lllcl}
    \toprule
    Data   Source          & Split                    & Task                                   & Number (k)    & Note                          \\ \midrule
    Open-GPT-4o-Image~\cite{chen2025opengpt}      & Generation               & Text-to-Image                          & 39k            & Generated by GPT-4o           \\
    BLIP3o~\cite{chen2025blip3}                 & Generation               & Text-to-Image                          & 60k            & Generated by GPT-4o       \\
    TextAtlas5M~\cite{wang2025textatlas5m}  & TextScenesHQ             & Text-to-Image                          & 36k            & Text rendering                \\
    TextAtlas5M~\cite{wang2025textatlas5m}  & Cover Book               & Text-to-Image                          & 207k           & Text rendering                \\
    TextAtlas5M~\cite{wang2025textatlas5m}   & LongWordsSubset-A        & Text-to-Image                          & 266k           & Text rendering                \\
    UniWorld-V1~\cite{lin2025uniworld}            & osp                      & Text-to-Image                          & 286k           & -                             \\
    ShareGPT-4o-Image~\cite{chen2025sharegpt4oimage}      & Generation               & Text-to-Image                          & 45k            & Generated by GPT-4o           \\ \midrule
    \textbf{Sum}           &                          & \textbf{Text-to-Image}                 & \textbf{939k}  &                               \\ \midrule
    OmniGen~\cite{omnigen}                & Subject-driven           & Image Editing                          & 192k           & Multi-reference generation    \\
    GPT-IMAGE-EDITE-1.5M~\cite{wang2025gpt}          & -                        & Image Editing                          & 1500k          & -                             \\
    Open-GPT-4o-Image~\cite{chen2025opengpt}      & Editing                  & Image Editing                          & 41k            & -                             \\
    ShareGPT-4o-Image~\cite{chen2025sharegpt4oimage}      & Editing                  & Image Editing                          & 46k            & -                             \\
    OmniGen2~\cite{wu2025omnigen2}               & Video Edit               & Image Editing                          & 1083k          & From video sequence           \\
    OmniGen2~\cite{wu2025omnigen2}               & Video ICEdit             & Image Editing                          & 155k           & From video sequence           \\
    OmniGen2~\cite{wu2025omnigen2}               & Video ICGen              & Image Editing                          & 888k           & From video sequence           \\
    SEED-Data-Edit~\cite{ge2024seed}               & -                        & Image Editing                          & 77k            & -                             \\
    StyleBooth~\cite{han2025stylebooth}             & -                        & Image Editing                          & 11k            & -                             \\
    NHR-Edit~\cite{kuprashevich2025nohumansrequired}               & -                        & Image Editing                          & 358k           & Generated by a in-house model \\
    Echo-4o-Image~\cite{ye2025echo}          & -                        & Image Editing                          & 73k            & -                             \\
    Pico-Banana-400k~\cite{qian2025pico}       & -                        & Image Editing                          & 400k           & -                             \\
    NanoConsistent~\cite{ye2025echo}         & -                        & Image Editing                          & 150k           & -                             \\
    ImgEdit~\cite{ye2025imgedit}                & -                        & Image Editing                          & 184k           & Filtered by VLM               \\
    UniWorld-V1~\cite{lin2025uniworld}            & Ghibli                   & Image Editing                          & 36k            & -                             \\
    UniWorld-V1~\cite{lin2025uniworld}            & ip\_img                  & Image Editing                          & 24k            & -                             \\
    UniWorld-V1~\cite{lin2025uniworld}            & omiedit                  & Image Editing                          & 368k           & -                             \\ \midrule
    \textbf{Sum}           & \textbf{}                & \textbf{Image Editing}                 & \textbf{5586k} &                               \\ \midrule
    OmniGen2~\cite{wu2025omnigen2}  & Video Interleaved        & Interleaved Context Alignment          & 657k           & -                             \\
    Video Dense Caption    & -                        & Interleaved Context Alignment          & 5000k          & Introduced in Sec.~\ref{sec:interleaved-alignment-data}         \\ \midrule
    \textbf{Sum}           & \textbf{}                & \textbf{Interleaved Context Alignment} & \textbf{5657k} & \textbf{}                     \\ \midrule
    Interleaved Generation & robot planning                  & Interleaved Generation                 & 168k            & Introduced in Sec.~\ref{sec:interleaved-instruction-tuning-data}         \\
    Interleaved Generation & cooking                  & Interleaved Generation                 & 15k            & Introduced in Sec.~\ref{sec:interleaved-instruction-tuning-data}         \\
    Interleaved Generation & general VQA              & Interleaved Generation                 & 15k            & Introduced in Sec.~\ref{sec:interleaved-instruction-tuning-data}         \\
    Interleaved Generation & web pages & Interleaved Generation                 & 268k           & Introduced in Sec.~\ref{sec:interleaved-instruction-tuning-data}         \\ \midrule
    \textbf{Sum}           & \textbf{}                & \textbf{Interleaved Generation}        & \textbf{466k}  & \textbf{}                     \\ \midrule
    \end{tabular}
    \caption{\textbf{Training data of the interleaved context alignment stage}.}
    \label{supp-tab:interleaved-context-alignment-data}
\end{table*}
\clearpage

\end{document}